\pdfoutput=1

\documentclass[11pt]{article}

\usepackage{amsmath}
\usepackage{bm}
\usepackage{xcolor}
\usepackage{caption}

\usepackage{algorithm}  
\usepackage{algpseudocode} 

\usepackage{float}
\usepackage[most]{tcolorbox}
\usepackage{enumitem} 
\usepackage{xcolor}

\usepackage[final]{acl}
\usepackage{multirow}
\usepackage{times}
\usepackage{latexsym}

\usepackage[T1]{fontenc}

\usepackage[utf8]{inputenc}

\usepackage{microtype}

\usepackage{inconsolata}

\usepackage{graphicx}
\usepackage{colortbl}
\usepackage{adjustbox}
\usepackage{xcolor}
\usepackage{amssymb}
\usepackage{amsfonts}
\usepackage{xspace}
\usepackage{amsmath}
\usepackage{booktabs}
\usepackage{graphicx} 
\newcommand{\model}{DPEPO\xspace}
\newcommand{\fullmodel}{Diverse Parallel Exploration Policy Optimization\xspace}

\newcommand{\ignore}[1]{}

\title{\model: Diverse Parallel Exploration Policy Optimization \\ for LLM-based Agents} 

\author{
\textbf{Junshuo Zhang$^{1,2}$, Chengrui Huang$^{1,2}$, Feng Guo$^1$, Zihan Li$^1$, Ke Shi$^1$} \\
\textbf{Menghua Jiang$^2$, Jiguo Yu$^1$, Shuo Shang$^1$ and Shen Gao$^1$} \\
$^1$University of Electronic Science and Technology of China, $^2$DiDi
}

\begin{document}
\maketitle
\begin{abstract} 
Large language model (LLM) agents that follow the sequential ``reason-then-act'' paradigm have achieved superior performance in many complex tasks.
However, these methods suffer from limited exploration and incomplete environmental understanding, as they interact with only a single environment per step. 
In this paper, we first introduce a novel paradigm that enables an agent to interact with multiple environments simultaneously and share cross-trajectory experiences. 
Building upon this paradigm, we further propose \fullmodel (\model), a reinforcement learning (RL) algorithm that encourages the agent to perform diverse parallel exploration.
There are two stages in \model: initial supervised fine-tuning (SFT) imparts basic parallel reasoning and action generation, followed by reinforcement learning stage with a hierarchical reward scheme. 
We design a parallel trajectory-level success reward and two step-level rewards: Diverse Action Reward and Diverse State Transition Reward, which actively penalize behavioral redundancy and promote broad exploration.
Extensive experiments on ALFWorld and ScienceWorld show that \model achieves state-of-the-art (SOTA) success rates, while maintaining comparable efficiency to strong sequential baselines.\footnote{Code is available at \url{https://github.com/LePanda026/Code-for-DPEPO}}
\end{abstract}

\section{Introduction}~\label{sec:intro}

\begin{figure}[t] 
\centering
  \includegraphics[width=\linewidth]{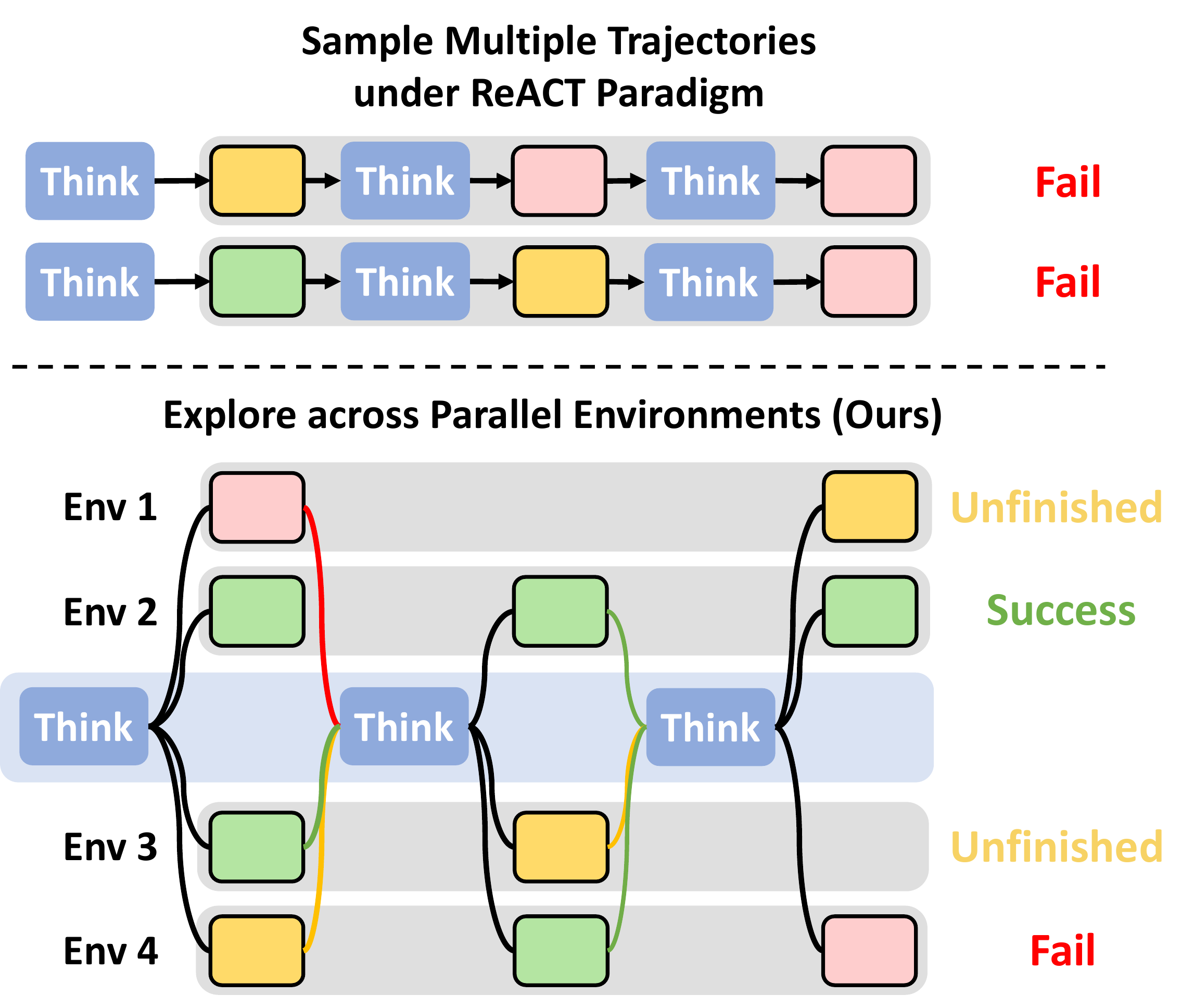} 
  \caption{ReAct-based agent constructs environmental understandings from a narrow, linear view. Our \model allows agent to build comprehensive cognition of environments from diverse parallel exploration.}
\end{figure} 

LLM-based autonomous agents commonly adopt the ReAct paradigm~\cite{react}, where the agent alternates between reasoning and acting: it selects a single action, observes the environmental feedback, and then uses that observation to inform its next step. 
This ``reason, then act'' framework has shown notable success in complex, long-horizon tasks~\cite{alfworld,webshop,scienceworld} by enabling structured, step-by-step reasoning. 
Recent work~\cite{agent_contain_world_model} demonstrates that a comprehensive understanding of the environment is crucial for effective and reliable agents. 
However, ReAct imposes a fundamental limitation: at each step, the agent can perform only one action and observe a single observation. 
This restricts the breadth of environmental exploration and forces the agent to construct its understanding from a narrow, linear view. 
As a result, agents often develop an incomplete and biased cognition of the environment, leading to suboptimal decisions and failures in complex tasks.


A straightforward approach to broaden exploration is to sample multiple independent trajectories for the same task. 
However, this naive multi-sampling strategy faces two critical shortcomings:
(1) \textbf{Lack of diversity}: Despite increased output entropy, actions sampled across multiple trials tend to converge to similar choices, failing to promote truly diverse exploration and preventing the agent from developing a comprehensive grasp of the environment.
(2) \textbf{Inefficiency and isolation}: Experiences from different trajectories are neither shared nor leveraged, meaning the agent cannot learn from cross-trajectory insights. 
Moreover, sampling multiple trials sequentially inflates token consumption and running time, making the approach prohibitively costly for practical deployment.


To overcome these limitations, we introduce parallel exploration, a paradigm in which an agent interacts with multiple environments simultaneously. 
This allows the agent to gather diverse observations in parallel, share experiences across environments, and build a more holistic understanding of the environment dynamics. 
Building on this exploration paradigm, we propose \fullmodel (\model), a novel reinforcement learning framework that endows agents with structured parallel exploration capabilities. 
After an initial SFT phase to acquire basic parallel reasoning and action generation skills, the agent is optimized via a hierarchical reward scheme that encourages diverse exploration at both the trajectory and step levels. 
Specifically, we design a parallel trajectory-level success reward and two novel step-level rewards: Diverse Action Reward and Diverse State Transition Reward, that actively penalize redundant behaviors. 
These rewards are combined through a group-relative advantage computation to provide multi-granularity learning signals, enabling the agent to explore more broadly while staying focused on task completion.
Extensive experiments on ALFWorld and ScienceWorld, demonstrate that \model achieves state-of-the-art success rates across all task splits. Notably, our analysis of trajectory length and token usage reveals that \model achieves efficiency comparable to strong sequential baselines, despite its parallel nature.


\noindent Our contributions are summarized as follows:




\noindent $\bullet$ We propose \fullmodel (\model), a novel RL framework that pioneers parallel exploration for agents to build holistic environment cognition.

\noindent $\bullet$ We design a Parallel Trajectory-level Success Reward to optimize task completion in concurrent multi-environment settings.

\noindent $\bullet$ We propose Parallel Step-level Rewards (Diverse Action \& State Transition) to penalize redundancy and foster distinct explorations.

\noindent $\bullet$ Extensive experiments on ALFWorld and ScienceWorld show \model achieves SOTA performance in terms of success rates while maintaining comparable efficiency with single exploration methods.
\section{Related Work}~\label{sec:related}

\paragraph{LLM as Autonomous Agents}
LLMs are increasingly being deployed as autonomous agents across a wide range of domains, including code generation \cite{codeagent}, web navigation \cite{webnavigation}, smart device control \cite{mcoaa,real_web_agent,infiguir1,mobile_agent_v3}, external tool use \cite{Confucius,cooperativeandinteractiveagents,toolleanringinthewild,ttpa} and interactive gaming \cite{alfworld,webshop,scienceworld,voyaga}. Early LLM-based agents typically relied on engineered prompting strategies \cite{cot} to achieve strong performance, such as ReAct \cite{react} and Reflexion \cite{reflection}. Recent studies has shifted their focus toward fine-tuning LLMs for agent-related tasks using either SFT \cite{eto,agentgym} or RL \cite{GiGPO,RLVMR}, aiming to develop agents with greater generalization ability and robustness, ultimately resulting in more efficient and resilient behaviors compared to earlier prompt-based methods.


\begin{figure*} 
  \centering
  \includegraphics[width=1.0\textwidth]{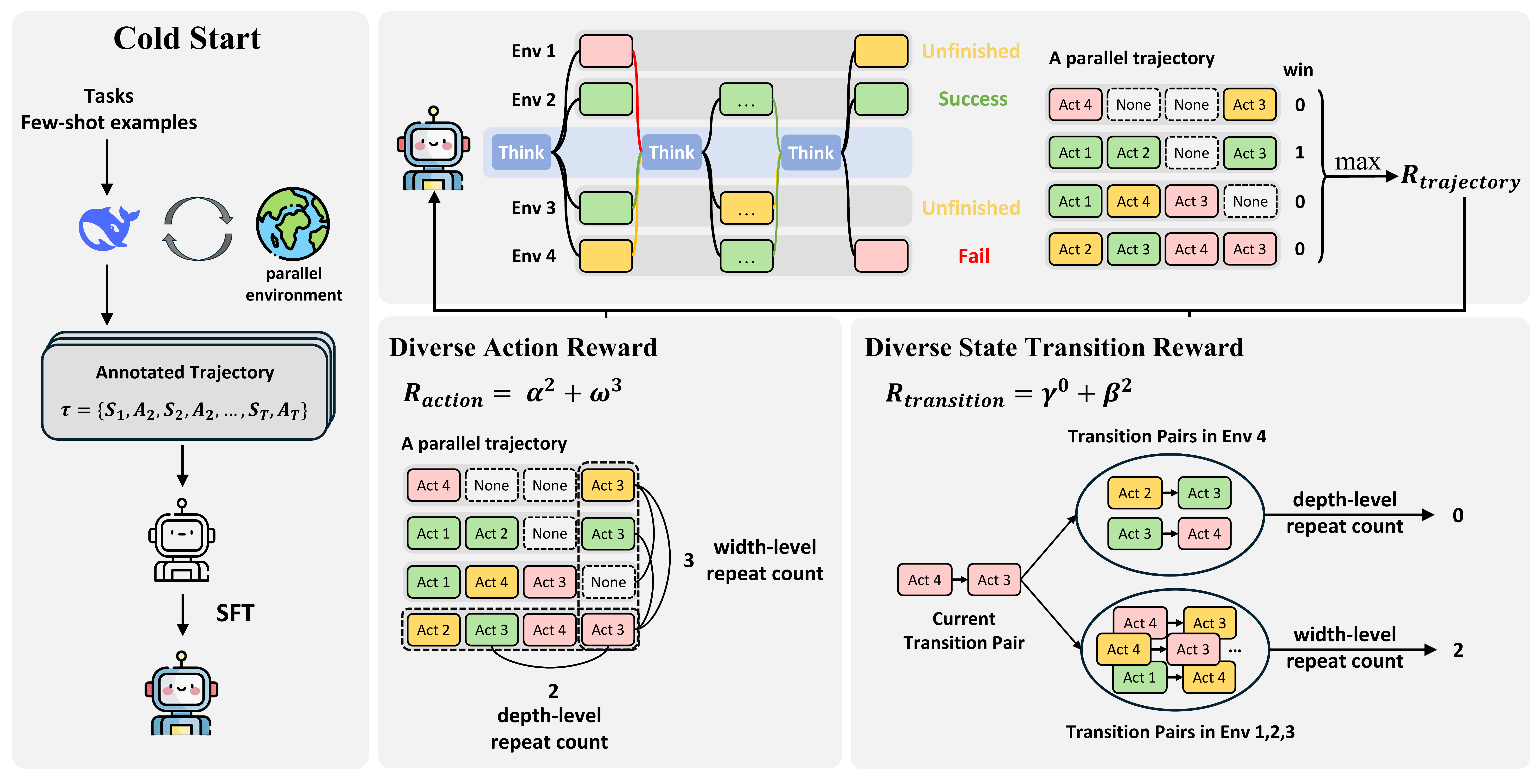} 
  \caption{Training framework of our proposed \fullmodel (\model).}
  \label{fig:model}
\end{figure*}

\paragraph{Reinforcement Learning for LLM-based Agents}
Reinforcement learning has been recognized as a foundational approach for enabling LLMs to operate as autonomous agents in complex environments. Early work in this domain primarily employed PPO \cite{ppo}, which allowed agents to learn diverse interactive tasks while avoiding the need for expensive imitation data.
With the emergence of reinforcement learning with verifiable reward (RLVR) \cite{deepseek-r1,tuluv3} and group-based policy optimization \cite{grpo,gspo,rloo,dr_grpo}, a series of studies have eliminated complex reward model and critic model, instead adopting rule-based reward signals: GiGPO \cite{GiGPO} proposes anchor state groups to estimate the quality of specific steps; RLVMR \cite{RLVMR} introduces meta-reasoning and corresponding step-level  rewards to address inefficient exploration; iStar \cite{istar} estimates step rewards via the LLM's intrinsic implicit rewards. However, these methods remain constrained by the sequential ReAct paradigm, which limits their exploration space and maintains a narrow focus on single-path reasoning. 
In this work, we introduce a novel reinforcement learning algorithm for training agents with more fine-grained reward shaping, drawing from the insight that more diverse and comprehensive exploration, helping agents build a better understanding of environments, thereby enabling them to perform tasks more effectively.

\section{Methodology}~\label{sec:method}

In this section, we first introduce the parallel agent exploration framework, and then we detail the framework for training the LLM-based agent to perform parallel exploration.
The overall architecture is shown in Figure~\ref{fig:model}.

\subsection{Parallel Agent Exploration Formulation}~\label{sec:problem_def}

Given a task specified by a prompt and a set of $K$ parallel environments $\mathcal{E} = \{E_1, E_2, \dots, E_K\}$ sharing the same initial state, an LLM-based agent interacts with $\mathcal{E}$ over multiple turns until the task is completed or the maximum allowed step $T$ is reached.
Specifically, at step $t$, the agent selects a subset of all environments $\mathcal{E}_t'$. Let $|\mathcal{E}_t'|$ denote the number of selected subset, due to the agent dynamically selects environments to interact, $|\mathcal{E}_t'|$ can vary at each step. 
For the selected environments, the agent generates a set of parallel actions $A_t = \{(E_i,a_t) \mid E_i \in \mathcal{E}_t' \}$, which are executed simultaneously in their respective environments. The agent then observes the resulting states $S_t = \{(E_i,s_t) \mid E_i \in \mathcal{E}_t' \}$.
Consequently, the agent's full trajectory can be denoted as:
\begin{equation} 
\tau = \{ (S_t, A_t) \big\}_{t=1}^{T},
\end{equation}
where the trajectory $\tau$ consists a sequence of parallel states and parallel actions.

\subsection{Parallel Exploration Capability Cold Start}\label{sec:coldstart}
We begin with an SFT-based cold start stage to endow the agent with foundational skills for structured output and parallel exploration. 
For the SFT training data, we introduce human prior knowledge into its synthetic pipeline. 
The data synthesis pipeline is two steps:
(1) We first manually annotated five ground-truth trajectories. Each trajectory consists of multiple steps, with some key steps containing two or more parallel exploration actions.
(2) Based on these trajectories, we employ DeepSeek-V3.2 to annotate trajectories containing parallel exploration behaviors.
Finally, the agent is fine-tuned on these annotated trajectories, learning to imitate the parallel exploration patterns.

Next, we employ a reinforcement learning method to train the LLM to perform better parallel exploration actions by using dual-level rewards as the training signal: trajectory-level and step-level.

\subsection{Parallel Trajectory-level Success Reward} 
In this section, we extend the rule-based success reward to parallel setting. Formally, given a task, parallel environments $\mathcal{E}$, agent interaction trajectory $\tau$ and task success state $\mathcal{G}$, the parallel trajectory-level success reward is calculated as follows:
\begin{equation} 
    R_{traj}(\tau) = 
    \begin{cases}
    1 & \text{if } S_T \cap \mathcal{G} \neq \emptyset, \\
    0, & \text{otherwise}.
    \end{cases}
\end{equation} 
This parallel trajectory-level success reward signifies that when any state in $S_T$ reaches a goal state, the overall trajectory is considered successful. This reward provides an important training signal that guides the agent to complete the task.

\subsection{Parallel Step-level Rewards}
The trajectory-level reward provides an important yet coarse-grained signal, failing to assess individual steps with parallel actions.
To provide fine-grained guidance for diverse exploration and better information acquisition, we introduce two novel step-level and diversity-driven rewards: \textbf{Diverse Action Reward} and \textbf{Diverse State Transition Reward}. 
These rewards discourage redundant exploration behavior, forcing the agent to gather more information for making better decisions. 
These dense, step-level reward signals complement the sparse trajectory-level reward, together providing a multi-granularity learning signal for parallel agent.

\paragraph{Diverse Action Reward} To incentivize diverse exploration behavior, we define a composite reward that incorporates both depth-wise (within a single environment) and width-wise (across different environments at the same step) considerations. 

Formally, at step \( t \), given the selected environments $\mathcal{E}_t'$ and generated parallel actions $A_t$.
We compute this reward as follows:
\begin{equation} 
R_{\text{action}}(A_t) = \frac{1}{|\mathcal{E}_t'|} \sum_{E_i \in \mathcal{E}t'} \alpha^{\,C_{\text{depth}}(E_i,a_t)} + \omega^{\,C_{\text{width}}(A_t)},
\label{eq:action_reward}
\end{equation} 
where $C_{\text{depth}}(E_i,a_t)$ is the number of repetitions of action $a_t$ in environment $E_i$, $\omega^{\,C_{\text{width}}(A_t)}$ is the number of repeated actions in $\mathcal{A}_t$. The \( \alpha, \omega \in (0, 1] \) are discount factors corresponding to depth-wise and width-wise rewards.
The Diverse Action Reward encourages diversity within each environment and between parallel environments.

\paragraph{Diverse State Transition Reward} 
Beyond individual actions, we further propose the diverse state transition reward to discourage the agent from repeating the same state-action transitions, this reward also operates at both depth and width level.

At step $t$, for each environment in selected $\mathcal{E}_t'$, the agent observes state $s_{t}$, takes an action $a_{t}$, define this state-action transition as $p_t = s_t \rightarrow a_t$. The parallel state $S_t$ and parallel actions $A_t$ result in parallel state-action transition $\{(E_i,p_t)|E_i \in \mathcal{E}_t'\}$.
We compute this reward as follows:
\begin{equation}
    \begin{split}
    R_{\text{transition}}(A_t) = &  \frac{1}{|\mathcal{E}_t'|} \sum_{E_i \in \mathcal{E}_t'}  \gamma^{M_{\text{depth}}(E_i,p_t)} \\
    \quad +  & \frac{1}{|\mathcal{E}_t'|} \sum_{E_i \in \mathcal{E}_t'}  \beta^{M_{\text{width}} (E_i,p_t)},
    \end{split}
\label{eq:transition_reward}
\end{equation}
where $M_{\text{depth}}(E_i,p_t)$ measures the recurrence of the transition \( p_t \) within the environment $E_i$, and $M_{\text{width}}(E_i,p_t)$ counts how many other selected environments have previously observed the same transition. The $\gamma, \beta \in (0, 1]$ are discount factors for the depth-wise and width-wise terms, respectively. 


\paragraph{Step-Level Composite Reward} The final reward for step $A_t$ is the average of \textbf{Diverse Action Reward} and \textbf{Diverse State Transition Reward}: 
\begin{equation}
    R_{step}(A_t) = \frac{R_{action}(A_t) + R_{transition}(A_t)}{2}.
\end{equation}
This reward provides a dense, multi-granularity policy optimizing signal that actively promotes diverse and efficient exploration.

\subsection{Policy Optimization}
Similar to GRPO~\cite{grpo}, \model eliminates the critic model and instead estimates the advantages of trajectories through group sampling and relative advantage computation. 
For a given task, we sample a group \( G \) containing \( N \) trajectories $\{\tau\}_{i=1}^N$, each trajectory paired with its corresponding trajectory reward $R_{traj}(\tau_i)$:
\begin{equation}
    G = \big\{ (\tau_i, R_{traj}(\tau_i)) \big\}_{i=1}^N.
\end{equation}
We then access the advantage of each trajectory by normalizing its reward using group-level statistics:  
\begin{equation}
    \Phi_{traj}(\tau_{i}) = \frac{R(\tau_{i}) - \text{mean}\left( \{ R_{traj}(\tau_{j}) \}_{j=1}^N \right)}{\text{std}\left( \{ R_{traj}(\tau_{j}) \}_{j=1}^N \right)}.
\label{eq:trajectory_advantage}
\end{equation}
The trajectory-level advantage $\Phi_{traj}(\tau_{i})$ reflects the agent’s overall success in completing the task. 
Similar to standard GRPO, it performs on outcome-level credit assignment, encouraging policy to develop long-horizon behaviors to complete the task. 
We use this trajectory-level advantage as the base advantage for more fine-grained, step-level advantage computation. 
More specifically, at step $t$, given a step $A_{i,t}$ in trajectories $\tau_i$ with its associated step reward $R_{\text{step}}(A_{i,t})$, we compute the step-level advantage by combining this immediate reward with the global trajectory advantage:
\begin{equation} 
\begin{small}
    \Phi_{step}(A_{i,t}) = 
    \begin{cases}
    R_{\text{step}}(A_{i,t}), & \text{if } \Phi_{traj}(\tau_{i}) > 0, \\
    2 - R_{\text{step}}(A_{i,t}), & \text{otherwise}.
    \end{cases}
\end{small}
\label{eq:step_advantage}
\end{equation}
\begin{equation}
     \Phi(A_{i,t}) = \Phi_{step}(A_{i,t}) \cdot \Phi_{traj}(\tau_i).
\label{eq:final_advantage}
\end{equation}
This combined advantage provides hierarchical supervision by assigning credit at both trajectory and step scales, thereby enabling the learned policies to exhibit more diverse behaviors across parallel environments.
We then optimize the policy for each step using the same objective as GRPO. 
The pseudo code for our algorithm can be found in Appendix~\ref{sec:pseudo_code}.
\begin{table*}[t]
\centering
\resizebox{\textwidth}{!}{
\begin{tabular}{lccccccccc|cccc}
\toprule
\multirow{2}{*}{Method} & \multicolumn{9}{c}{\textbf{ALFWorld}} & \multicolumn{4}{c}{\textbf{ScienceWorld}} \\  
& Pick & Cool & Heat & Look & Pick2 & Clean & All & OOD & Avg. & L0 & L1 & L2 & Avg. \\
\midrule
\multicolumn{13}{l}{\textit{Frontier LLMs}} \\
GPT-4o & 75.3 & 21.6& 56.7& 60.8& 49.8& 31.2 &48.0 & 66.0 & 57.0 & 45.4 &49.2 &41.0 & 45.2  \\
DeepSeek-V3 & 71.4 & 76.0& 81.3& 61.5& 75.0 & 77.8& 74.3& 80.6& 77.5& 27.3& 35.2& 26.5& 29.7 \\
DeepSeek-R1 & 80.0 & 80.0& 56.3& 46.2& 87.5& 77.8& 75.0& 85.1& 80.1& 22.2& 31.4& 29.1& 27.6 \\

\midrule
\multicolumn{13}{l}{\textit{Qwen2.5-1.5B-Instruct}} \\
~~ETO  & - & - & - & - & - & - & 64.1 & 66.4 & 65.3 & 39.1 & 22.7 & 15.6 & 25.8 \\
~~GRPO &85.3 &59.7 &78.2 &53.7 &53.5 &84.5 &72.8 &71.1 &72.0 &21.1 &13.7 &10.9 &15.2 \\
~~GiGPO &94.4 &79.8 &\underline{94.4} &67.5 &76.4 &94.8 &86.7 &83.2 &85.0 &25.8 &15.2 &4.7 &15.2 \\
~~SPEAR &\underline{96.4} &\underline{87.6} &\textbf{99.0} &\underline{86.5} &\textbf{91.6} &\underline{96.1} &\underline{93.2} &- &- &- &- &- &- \\
~~RLVMR &- &- &- &- &- &- &89.1 &\underline{87.9} &\underline{88.5} &\underline{46.9} &\underline{34.3} &\underline{26.5} &\underline{35.9} \\
~~\model &\textbf{100.0} &\textbf{96.0} &93.8 &\textbf{100.0} &\underline{87.5} &\textbf{96.3} &\textbf{95.7} &\textbf{92.5} &\textbf{94.1} &\textbf{59.8} &\textbf{58.1} &\textbf{34.2} &\textbf{50.7} \\

\midrule
\multicolumn{13}{l}{\textit{Qwen2.5-7B-Instruct}} \\
~~ETO &- &- &- &- &- &- &70.3 &74.2 &72.3 &62.5 &40.6 &28.1 &43.7 \\
~~GRPO &90.8 &72.5 &74.7 &66.1 &77.6 &89.3 &77.6 &77.3 &77.5 &49.1 &30.1 &26.6 &35.3 \\
~~GiGPO &97.7 &89.3 &83.7 &82.7 &79.2 &\textbf{98.8} &90.8 &90.2 &90.5 &53.4 &25.2 &25.8 &34.8 \\ 
~~SPEAR &\underline{99.9} &\underline{89.9} &\textbf{96.4} &\underline{85.1} &\underline{95.1} &95.6 &\underline{94.7} &- &- &- &- &- &- \\
~~RLVMR &- &- &- &- &- &- &91.4 &\underline{91.8} &\underline{91.6} &\textbf{67.2} &\underline{43.0} &\underline{32.2} &\underline{47.5} \\
~~\model &\textbf{100.0} &\textbf{100.0} &\underline{93.8} &\textbf{100.0} &\textbf{100.0} &\underline{96.3} &\textbf{98.6} &\textbf{97.8} &\textbf{98.2} &\underline{66.6} &\textbf{66.5} &\textbf{51.0} &\textbf{61.4 }\\
\bottomrule
\end{tabular}
} 
\caption{Performance comparison on ALFWorld and ScienceWorld. For ALFWorld, we report success rate (\%) on seen (in-domain) and unseen (out-of-domain) tasks. For ScienceWorld, we report results on three splits: L0 (seen variants and categories), L1 (unseen variants, seen categories), and L2 (unseen variants and categories). \model achieves superior in-domain performance and strong generalization ability. \textbf{Bold} is best, \underline{underlined} is second-best.}
\label{tab:main_results}
\end{table*}




\section{Experimental Setup}
\subsection{Datasets} 
We train and evaluate the LLM-based agents on two challenging benchmarks: ALFWorld~\cite{alfworld} and ScienceWorld~\cite{scienceworld}. ALFWorld is a text-based interactive environment where agents perform everyday household tasks through multi-step reasoning and interactions. It presents goals like ``put the pencil on the desk'', requiring agents to explore simulated rooms based on common sense. The environment is fully text-based and only returns a reward signal upon final success or failure. ScienceWorld is a simulated benchmark designed to evaluate scientific agent's reasoning capabilities across diverse elementary science topics. It includes tasks from 10 domains, including electricity, life cycles, and chemical reactions. Each task consists of multiple intermediate sub-goals that agents must complete sequentially. 

\subsection{Baselines}
For ALFWorld and ScienceWorld, we compare our approach with a range of competitive baselines under the ReAct paradigm: (1) Closed-source LLMs: GPT-4o~\cite{gpt-4o}, DeepSeek-V3~\cite{deepseek_v3}, and DeepSeek-R1~\cite{deepseek-r1}, which show superior performance on various natural language tasks. (2) A series of RL methods: 
GRPO~\cite{grpo}, which performs advantage estimation over trajectory groups; GiGPO~\cite{GiGPO}, introducing step-level advantage through anchor state grouping; RLVMR~\cite{RLVMR}, which proposes meta-reasoning rewards to address inefficient exploration; and SPEAR~\cite{spear}, leveraging self-imitation and intrinsic rewards to improve the exploration–exploitation balance.

\subsection{Implementation Details}
We conduct experiments on various model sizes of Qwen2.5-Instruct~\cite{qwen25} and Qwen3~\cite{qwen3}. For cold start training, we annotate approximately 500 and 1,000 for ALFWorld and ScienceWorld, respectively. For reinforcement learning stage, the number of training sample is 500, the temperature for rollout is 0.4, the rollout group size $N$ is set to 8 for ALFWorld and 4 for ScienceWorld, with 1 epoch and a total of 125 training steps. The max step is 25, and the number of parallel environments is 4. Full hyperparameter details and prompts used for training are provided in Appendix ~\ref{app:experiment_details} and Appendix ~\ref{app:prompts}.

\section{Experimental Results}

\subsection{Overall Performance}
Table~\ref{tab:main_results} demonstrates the performance of \model and baselines on ALFWorld and ScienceWorld datasets. 
As illustrated, closed-source LLMs exhibit only limited performance: On ALFWorld, DeepSeek-R1 achieves a success rate of 80.1\%, and on ScienceWorld, GPT-4o reaches 45.2\%. 
Under the ReACT paradigm, RL training yields substantial improvements: GRPO raises the average success rate on ALFWorld to 72.0\% for the 1.5B model and 77.5\% for the 7B model, demonstrating the effectiveness of rule-based success reward. However, the coarse-grained nature of this reward limits further performance gains.
In contrast, GiGPO, SPEAR and RLVMR address this issue by incorporating step-level rewards that provide more fine-grained optimization signals for intermediate reasoning steps. Among these methods, RLVMR achieves the strongest performance, with average success rates of 91.6\% on ALFWorld and 47.5\% on ScienceWorld.
Despite the demonstrated effectiveness of these RL methods, they remain constrained by the  ReAct paradigm, which restricts the breadth of environmental exploration and forces the agent to construct its understanding from a narrow, linear view.
Our proposed \model addresses this limitation through parallel exploration and diversity-driven rewards, achieving the best results: 98.2\% on ALFWorld and 61.4\% on ScienceWorld. It also demonstrates strong generalization on unseen ALFWorld tasks and on the L1 and L2 splits of ScienceWorld.
This significant improvement can be attributed to diverse parallel exploration, which allows agents to gain a comprehensive understanding of the environment, thereby supporting more informed decision-making and ultimately leading to more successful task completion.

\begin{figure}
  \centering
  \includegraphics[width=\linewidth]{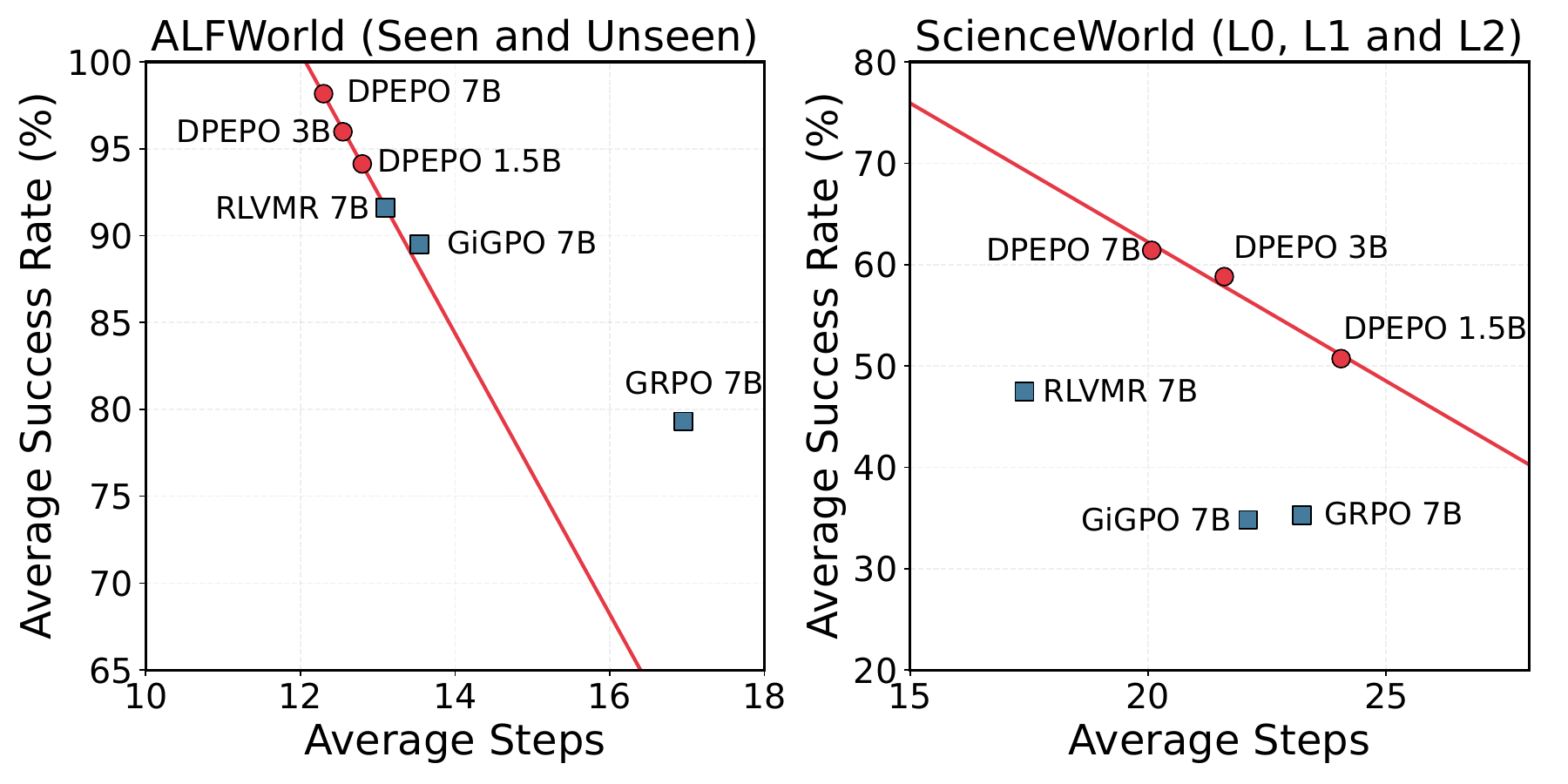} 
  \caption{Scaling experiments of \model with varying model sizes. The x-axis and y-axis show the average steps and average success rate across all splits.}
  \label{fig:length_analysis}
\end{figure}
\subsection{Scaling Abilities of \model}~\label{sec:scaling_abilities}
In this section, to demonstrate the scaling potential of \model, we present its performance across models with varying model sizes.
As shown in Figure~\ref{fig:length_analysis}, \model exhibits consistent performance improvements and higher efficiency as the model size increases, underscoring its generalization and scaling abilities.
Moreover, on the ALFWorld benchmark, even our smallest \model variant (based on Qwen2.5-1.5B-Instruct) outperforms the strong baseline RLVMR by 2.53\% in success rate and completes tasks in fewer steps.
On the more challenging ScienceWorld benchmark, \model further achieves higher performance with competitive efficiency. This highlights that diverse parallel exploration enables more effective and efficient task completion. More scaling experiments conducted on Qwen3 series can be found in Appendix~\ref{app:scaling_qwen3}.

\subsection{Analysis of Environment Numbers} 

\begin{figure}[ht] 
\centering
  \includegraphics[width=\linewidth]{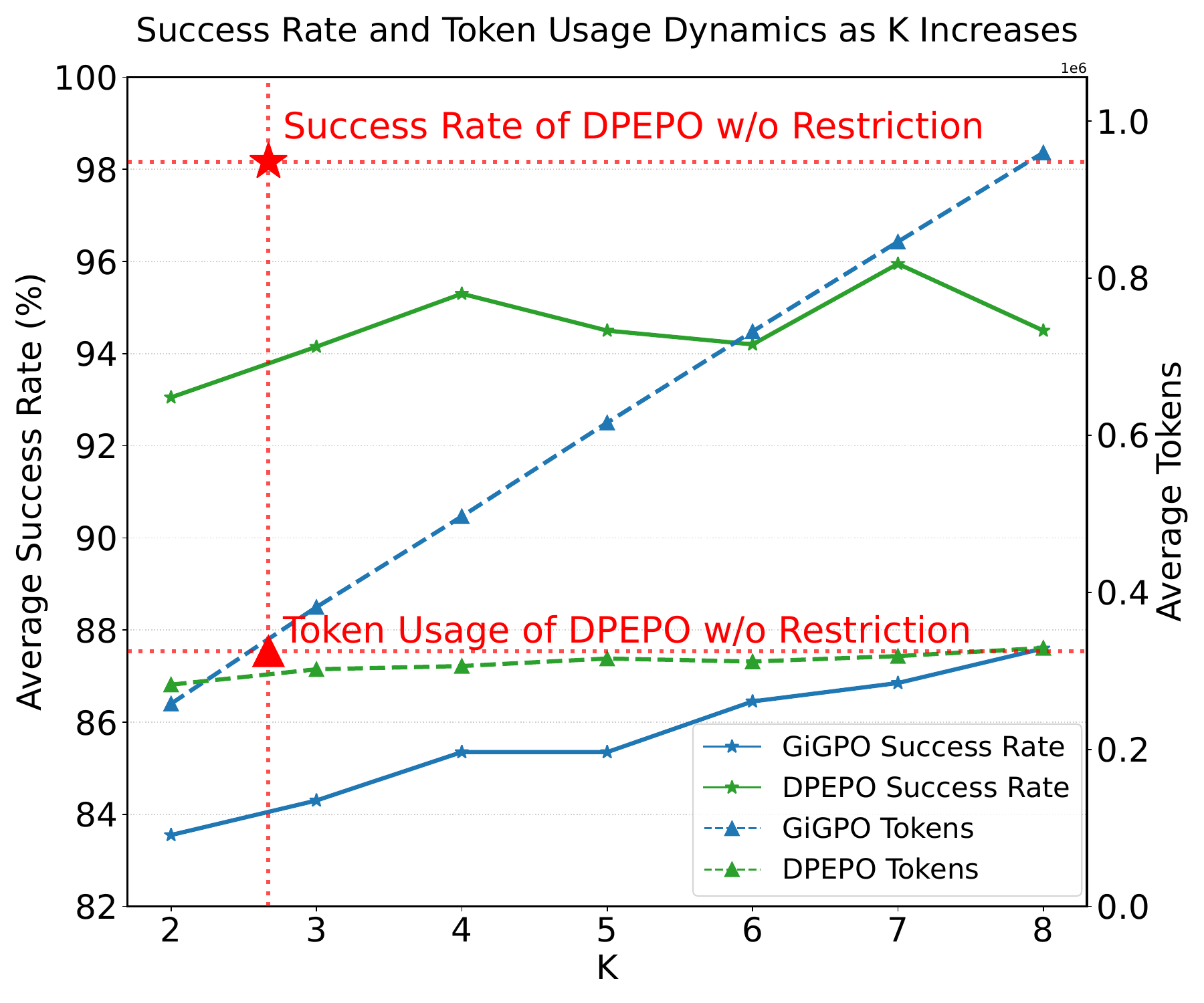} 
  \caption{Average success rate and token budgets on ALFWorld's seen and unseen splits, under two scaling strategies: (1) increasing sample count for GiGPO, and (2) increasing the number of environments for \model.}
  \label{fig:parallel_scaling}
\end{figure}


Furthermore, we analyze the impact of varying the number $K$ of parallel environments during inference. For this analysis, we evaluate on both seen and unseen tasks in ALFWorld and report the average success rate. During \model inference, we introduce a limiting prompt that specifies the maximum number of parallel environments the agent is allowed to explore; any trial that exceeds $K$ is treated as a failure. We use GiGPO with multiple sampling as our baseline; in GiGPO, $K$ denotes the number of sampled trajectories.

As shown in Figure~\ref{fig:parallel_scaling}, both \model and GiGPO exhibit stable performance improvement trends as $K$ increases, but \model consistently outperforms GiGPO across all values of $K$. Moreover, \model achieves strong performance even at small $K$, demonstrating high sample efficiency. 
In terms of token usage, the token budget of \model is only slightly higher than that of GiGPO when $K$ is small. As $K$ increases, GiGPO’s token consumption increases approximately linearly, while the token budget of \model grows much more slowly.
Notably, \model without any restriction on exploration environments not only achieves the best performance, but also explores fewer environments on average. The reason for this phenomenon is that the environmental limit prompt may introduce additional inductive bias, causing the model to explore all environments exhaustively rather than selectively.

\subsection{Ablation Study}\label{sec:ablation_study}

\begin{figure}[ht!] 
\centering
  \includegraphics[width=\linewidth]{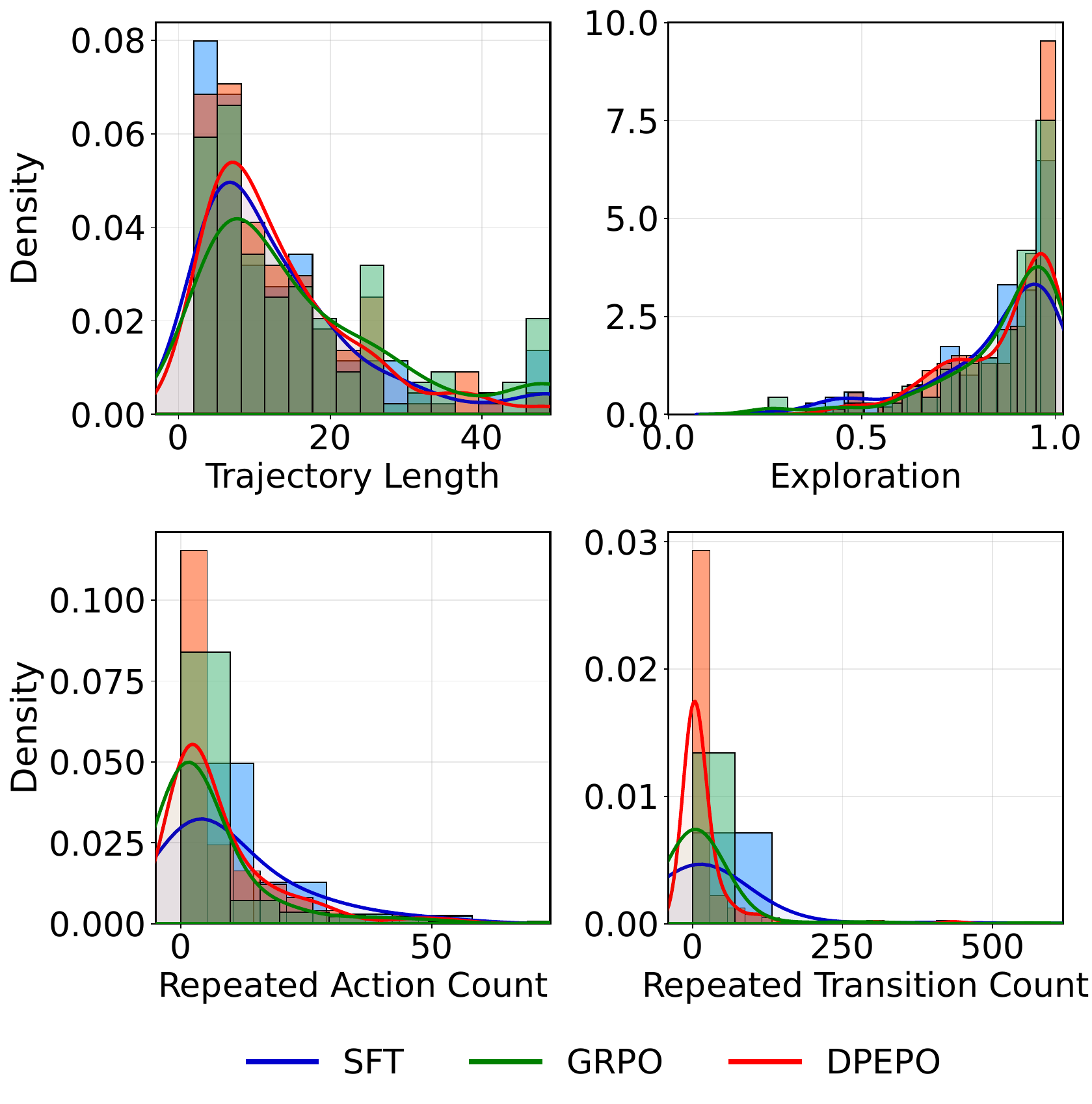} 
  \caption{Analysis of exploration dynamics across ablation variants on the ALFWorld seen-task split.}
  \label{fig:ablation_dynamics}
\end{figure}
Next, we conducted an ablation study to evaluate the contribution of each proposed reward component.
We compare \model against four variants: the variant only performed the cold start stage (ColdStart), \model without Diverse Action Reward (\model w/o DAR), \model without Diverse State Transition Reward (\model w/o DTR), and \model with neither reward (\model w/o DAR \& DTR). As shown in Table~\ref{tab:ablation_study}, the ColdStart variant yields only moderate performance. Removing either DAR or DTR leads to a significant performance degradation, and the variant without both rewards shows an even greater decline.
This suggests that DAR and DTR jointly provide training signals for the policy optimization process, thus contributing to the superior performance of \model. More detailed results of specific tasks can be found in Appendix~\ref{app:ablation_details}.

\begin{table}
\centering
\resizebox{\columnwidth}{!}{
\begin{tabular}{lcc|ccc}
\toprule
\multirow{2}{*}{Method} & \multicolumn{2}{c}{\textbf{ALFWorld}} & \multicolumn{3}{c}{\textbf{ScienceWorld}} \\  
& In Domain & OOD & L0 & L1 & L2 \\
\midrule 
ColdStart &93.6 &\underline{97.8} &66.5 &62.2 &48.1 \\
DPEPO &\textbf{98.6} &\underline{97.8} &\textbf{66.6} &\textbf{66.5} &\textbf{51.0} \\
~~w/o DAR &97.1 &97.0 &65.1 &62.8 &49.0 \\
~~w/o DTR &96.4 &98.5 &64.4 &63.6 &49.7 \\
~~w/o DAR \& DTR &\underline{96.4} &\textbf{98.5} &\underline{66.3} &\underline{65.7} &\underline{49.9} \\
\bottomrule
\end{tabular}
}
\caption{Ablation results on ALFWorld and ScienceWorld. \textbf{Bold scores} indicate the best performance, while \underline{underlined scores} represent the second-best.}
\label{tab:ablation_study}
\end{table}

To further understand how these reward components influence agent's behavior, we visualize several exploration metrics across aforementioned variants: (1) exploration diversity, (2) state transition repeat count, (3) action repeat count, and (4) trajectory length. The repetition counts are computed using Equations~\ref{eq:action_reward} and ~\ref{eq:transition_reward}, exploration diversity is the fraction of unique actions among all actions in parallel trajectories.
As illustrated in Figure~\ref{fig:ablation_dynamics}, agents trained with \model exhibit significantly lower action and transition repetition and higher exploration diversity compared to ablated variants.
This indicates that the DAR and DTR shape the agent's policy to generate trajectories that are not only more goal-directed but also more diverse and efficient, thereby achieving a better trade-off between exploration and task success.

\subsection{Efficiency Analysis}~\label{sec:explore_efficient_analysis}

\subsubsection{Inference-time Efficiency Analysis} 
Having established the effectiveness of our approach in terms of success rate, we now turn to its efficiency: while diverse parallel exploration behavior broadens the search space, it may also lead to a higher computational budget or inference time.

To quantitatively assess this trade-off, 
We compare the token budget and inference time of DPEPO with those of various methods on ALFWorld.
As shown in Table~\ref{tab:latency_analysis}, although understanding parallel environments brings higher token budgets, \model completes tasks with significantly shorter trajectories and faster inference time compared to GiGPO and the DeepSeek series.
We attribute this to two factors: (1) simultaneously executed parallel actions do not increase inference time, and (2) parallel interaction enables the agent to acquire more comprehensive environmental insights, supporting more purposeful decision-making that leads to more efficient and successful task completion.
\begin{figure}
  \centering
  \includegraphics[width=\linewidth]{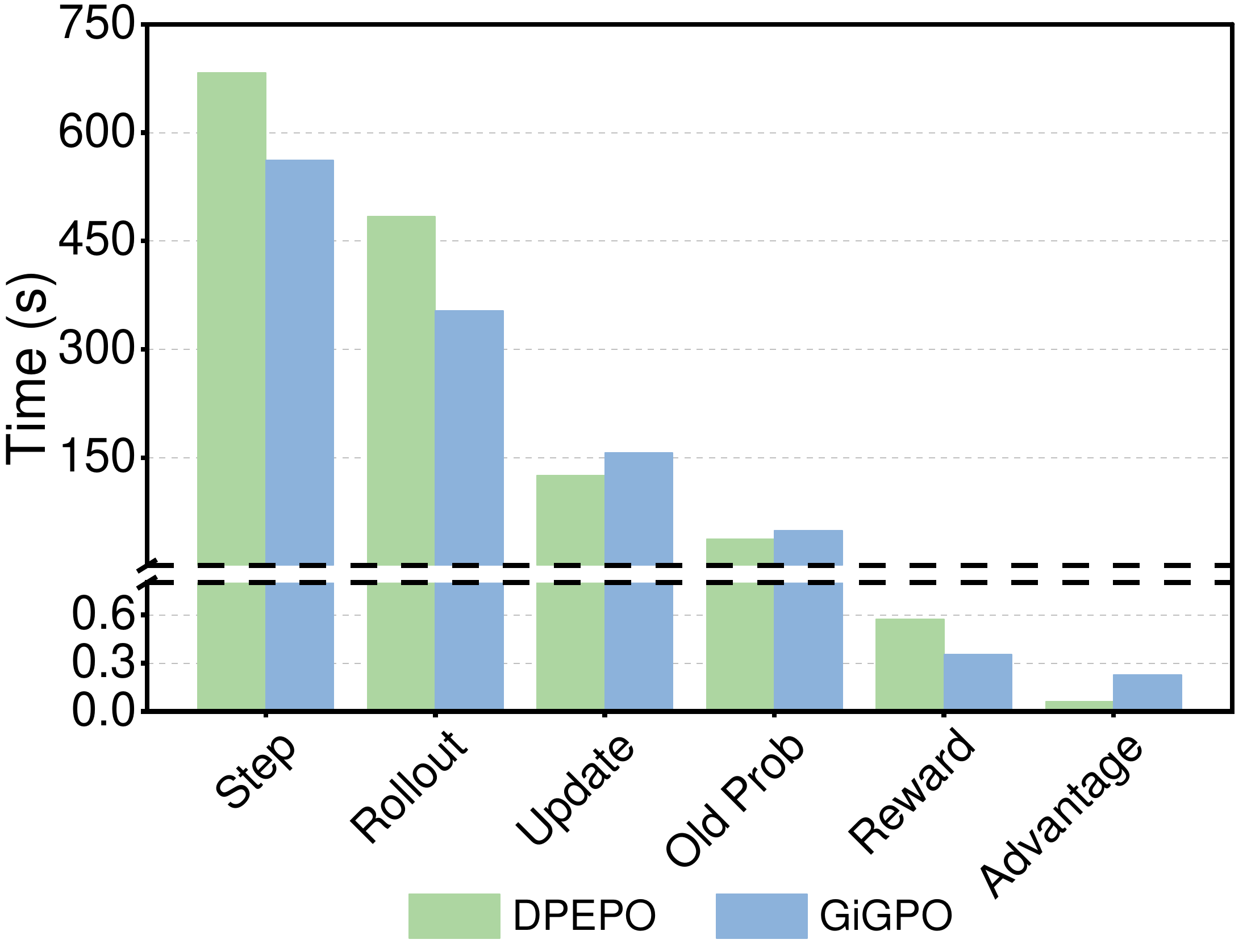} 
  \caption{Time consumption details of the main components at each step during \model training process.}
  \label{fig:time_details}
\end{figure}

\subsubsection{Training Efficiency Analysis}~\label{sec:train_efficiency}
\begin{figure} 
  \centering
  \includegraphics[width=\linewidth]{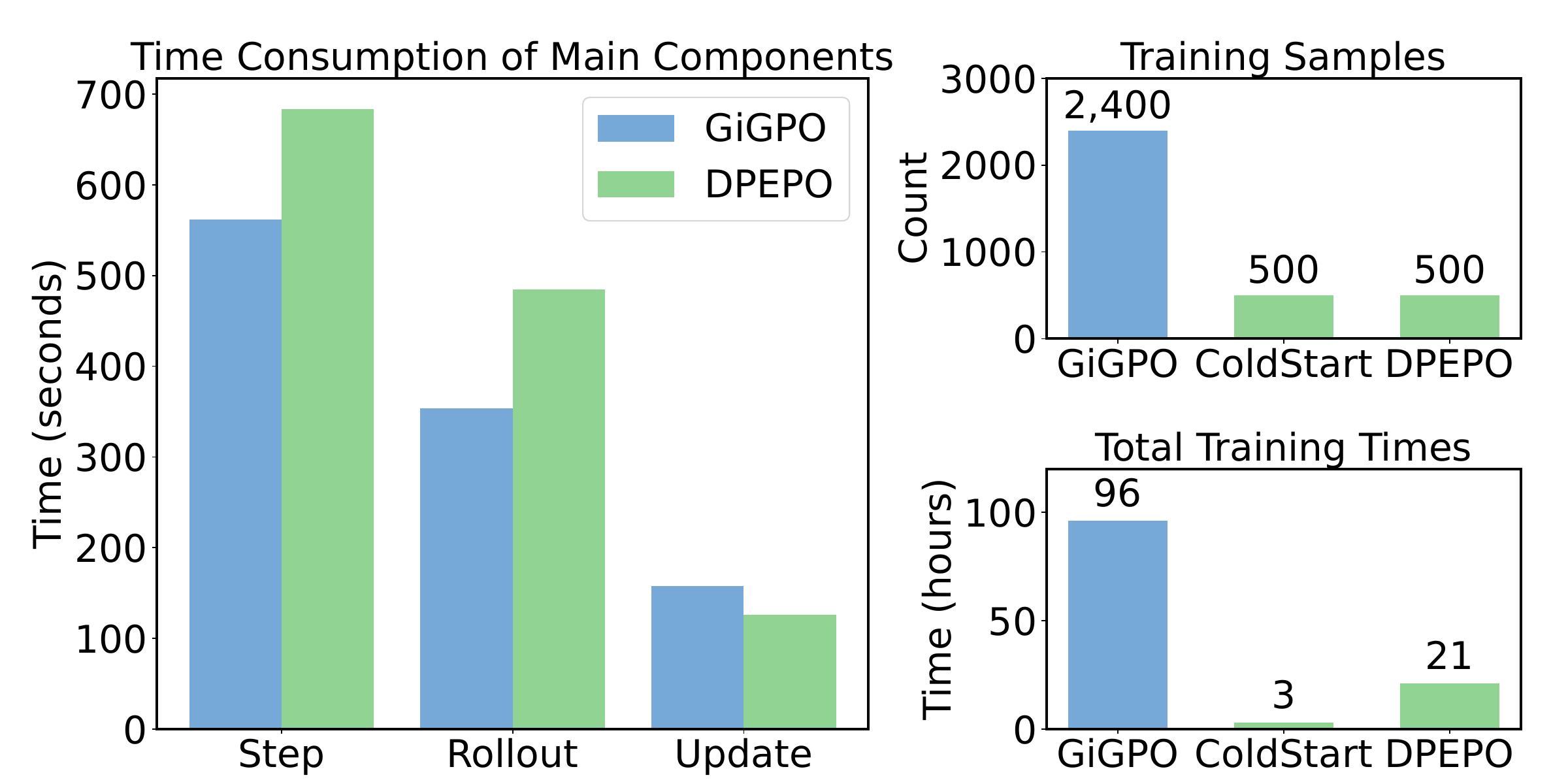} 
  \caption{Comparison of training efficiency.}
  \label{fig:train_efficiency}
\end{figure}

We further evaluate the training efficiency of \model in comparison with GiGPO, a critic-free method known for its training efficiency and strong performance. 
We first compare the time consumption per reasoning step. As shown in the left part of Figure~\ref{fig:time_details} and Figure~\ref{fig:train_efficiency}, although our method explores more environments in parallel, the time consumption per step does not increase significantly.
Despite the longer time per step, the overall training efficiency of \model is significantly higher: as shown in the right part of Figure~\ref{fig:train_efficiency}, including the cold start stage, \model achieves superior final performance in approximately 24 hours using only 500 RL training samples. In contrast, GiGPO requires 96 hours of training to reach its peak performance.
This experiment is conducted using Qwen2.5-1.5B-Instruct. 
Further details and experimental setup of this experiment are provided in Appendix~\ref{app:time_details}.
\begin{table}[t]
\centering
\begin{tabular}{lccccccc}
\toprule 
\multirow{1}{*}{\textbf{Method}} & \multirow{1}{*}{\textbf{Tokens}} & \multirow{1}{*}{\textbf{Steps}} & \multirow{1}{*}{\textbf{Time (s)}} \\  
\midrule 
DeepSeek-V3 &950.0 &20.5 & 62.4  \\
DeepSeek-R1 &1667.9 &24.8 & 237.0   \\
GiGPO       &1115.1 &15.2 &70.8  \\
DPEPO       &2283.4 &12.3 &44.7  \\
\bottomrule
\end{tabular}
\caption{Inference time comparison between \model and other methods under identical experimental settings.}
\label{tab:latency_analysis} 
\end{table}

\section{Conclusion}
In this work, we propose a novel parallel exploration paradigm and reinforcement learning algorithm to address the constrained exploration challenge inherent in ReACT paradigm. We first extend the ReACT framework to enable agents to simultaneously explore multiple parallel environments. We then introduce \fullmodel (\model), a novel diversity-driven policy gradient algorithm specifically designed for parallel agents. By incorporating fine-grained, step-level rewards into the parallel trajectory-level success reward, \model encourages agents to perform more diverse exploration in parallel environments. 
Empirical evaluations across all test splits on ALFWorld and the more challenging ScienceWorld benchmarks demonstrate that \model achieves state-of-the-art performance and high sample efficiency. Through extensive experiments and analysis, we show that the diverse parallel exploration behavior reinforced by \model leads to superior task performance, higher efficiency, and more effective token usage.

\clearpage
\section*{Acknowledgments}
This work was supported by the National Natural Science Foundation of China (62432002 and 62406061), the CCF-DiDi GAIA Collaborative Research Funds (CCF-DiDi GAIA 202504), and the State Key Laboratory of Internet Architecture, Tsinghua University (HLW2025MS10).

\section*{Limitations}

While \model shows significant improvements in performance and efficiency, it still faces challenges in real-world scenarios where parallel environments are hard to build.
However, we suggest a possible solution: training agents to complete different tasks in parallel, which might help mitigate this issue in real-world scenarios. We leave this exploration for future work. 

\section*{Ethical Considerations}

In some real-world environments that may contain biases, parallel exploration enables the agent to gather more experience, which could lead it to learn those biases, potentially resulting in unfair or unsafe behavior.
Therefore, before applying our proposed method to real-world tasks, developers should conduct thorough evaluations in their target environments to ensure the agent acquires accurate and objective knowledge and performs actions safely and responsibly.

\bibliography{custom}

@misc{alfworld,
      title={ALFWorld: Aligning Text and Embodied Environments for Interactive Learning}, 
      author={Mohit Shridhar and Xingdi Yuan and Marc-Alexandre Côté and Yonatan Bisk and Adam Trischler and Matthew Hausknecht},
      year={2021},
      eprint={2010.03768},
      archivePrefix={arXiv},
      primaryClass={cs.CL},
      url={https://arxiv.org/abs/2010.03768}, 
}

@misc{scienceworld,
      title={ScienceWorld: Is your Agent Smarter than a 5th Grader?}, 
      author={Ruoyao Wang and Peter Jansen and Marc-Alexandre Côté and Prithviraj Ammanabrolu},
      year={2022},
      eprint={2203.07540},
      archivePrefix={arXiv},
      primaryClass={cs.CL},
      url={https://arxiv.org/abs/2203.07540}, 
}

@inproceedings{agent_contain_world_model,
  title={General agents need world models},
  author={Richens, Jonathan and Everitt, Tom and Abel, David},
  booktitle={Forty-second International Conference on Machine Learning},
  year={2025}
}

@misc{GiGPO,
      title={Group-in-Group Policy Optimization for LLM Agent Training}, 
      author={Lang Feng and Zhenghai Xue and Tingcong Liu and Bo An},
      year={2025},
      eprint={2505.10978},
      archivePrefix={arXiv},
      primaryClass={cs.LG},
      url={https://arxiv.org/abs/2505.10978}, 
}

@misc{RLVMR,
      title={RLVMR: Reinforcement Learning with Verifiable Meta-Reasoning Rewards for Robust Long-Horizon Agents}, 
      author={Zijing Zhang and Ziyang Chen and Mingxiao Li and Zhaopeng Tu and Xiaolong Li},
      year={2025},
      eprint={2507.22844},
      archivePrefix={arXiv},
      primaryClass={cs.LG},
      url={https://arxiv.org/abs/2507.22844}, 
}

@inproceedings{codeagent,
    title = "{C}ode{A}gent: Enhancing Code Generation with Tool-Integrated Agent Systems for Real-World Repo-level Coding Challenges",
    author = "Zhang, Kechi  and
      Li, Jia  and
      Li, Ge  and
      Shi, Xianjie  and
      Jin, Zhi",
    editor = "Ku, Lun-Wei  and
      Martins, Andre  and
      Srikumar, Vivek",
    booktitle = "Proceedings of the 62nd Annual Meeting of the Association for Computational Linguistics (Volume 1: Long Papers)",
    month = aug,
    year = "2024",
    address = "Bangkok, Thailand",
    publisher = "Association for Computational Linguistics",
    url = "https://aclanthology.org/2024.acl-long.737/",
    doi = "10.18653/v1/2024.acl-long.737",
    pages = "13643--13658"
}

@misc{webnavigation,
      title={Agent-E: From Autonomous Web Navigation to Foundational Design Principles in Agentic Systems}, 
      author={Tamer Abuelsaad and Deepak Akkil and Prasenjit Dey and Ashish Jagmohan and Aditya Vempaty and Ravi Kokku},
      year={2024},
      eprint={2407.13032},
      archivePrefix={arXiv},
      primaryClass={cs.AI},
      url={https://arxiv.org/abs/2407.13032}, 
}

@inproceedings{mcoaa,
    title = "You Only Look at Screens: Multimodal Chain-of-Action Agents",
    author = "Zhang, Zhuosheng  and
      Zhang, Aston",
    editor = "Ku, Lun-Wei  and
      Martins, Andre  and
      Srikumar, Vivek",
    booktitle = "Findings of the Association for Computational Linguistics: ACL 2024",
    month = aug,
    year = "2024",
    address = "Bangkok, Thailand",
    publisher = "Association for Computational Linguistics",
    url = "https://aclanthology.org/2024.findings-acl.186/",
    doi = "10.18653/v1/2024.findings-acl.186",
    pages = "3132--3149"
}

@inproceedings{real_web_agent,
 author = {Gur, Izzeddin and Furuta, Hiroki and Huang, Austin and Safdari, Mustafa and Matsuo, Yutaka and Eck, Douglas and Faust, Aleksandra},
 booktitle = {International Conference on Representation Learning},
 editor = {B. Kim and Y. Yue and S. Chaudhuri and K. Fragkiadaki and M. Khan and Y. Sun},
 pages = {52690--52717},
 title = {A Real-World WebAgent with Planning, Long Context Understanding, and Program Synthesis},
 url = {https://proceedings.iclr.cc/paper_files/paper/2024/file/e91bf7dfba0477554994c6d64833e9d8-Paper-Conference.pdf},
 volume = {2024},
 year = {2024}
}

@misc{mobile_agent_v3,
      title={Mobile-Agent-v3: Fundamental Agents for GUI Automation}, 
      author={Jiabo Ye and Xi Zhang and Haiyang Xu and Haowei Liu and Junyang Wang and Zhaoqing Zhu and Ziwei Zheng and Feiyu Gao and Junjie Cao and Zhengxi Lu and Jitong Liao and Qi Zheng and Fei Huang and Jingren Zhou and Ming Yan},
      year={2025},
      eprint={2508.15144},
      archivePrefix={arXiv},
      primaryClass={cs.AI},
      url={https://arxiv.org/abs/2508.15144}, 
}

@misc{voyaga,
      title={Voyager: An Open-Ended Embodied Agent with Large Language Models}, 
      author={Guanzhi Wang and Yuqi Xie and Yunfan Jiang and Ajay Mandlekar and Chaowei Xiao and Yuke Zhu and Linxi Fan and Anima Anandkumar},
      year={2023},
      eprint={2305.16291},
      archivePrefix={arXiv},
      primaryClass={cs.AI},
      url={https://arxiv.org/abs/2305.16291}, 
}

@inproceedings{
react,
title={{ReAct}: Synergizing Reasoning and Acting in Language Models},
author={Shunyu Yao and Jeffrey Zhao and Dian Yu and Nan Du and Izhak Shafran and Karthik R Narasimhan and Yuan Cao},
booktitle={The Eleventh International Conference on Learning Representations },
year={2023},
url={https://openreview.net/forum?id=WE_vluYUL-X}
}

@article{reflection,
  title={Reflexion: Language agents with verbal reinforcement learning},
  author={Shinn, Noah and Cassano, Federico and Gopinath, Ashwin and Narasimhan, Karthik and Yao, Shunyu},
  journal={Advances in Neural Information Processing Systems},
  volume={36},
  pages={8634--8652},
  year={2023}
}

@article{eto,
  title={Trial and error: Exploration-based trajectory optimization for llm agents},
  author={Song, Yifan and Yin, Da and Yue, Xiang and Huang, Jie and Li, Sujian and Lin, Bill Yuchen},
  journal={arXiv preprint arXiv:2403.02502},
  year={2024}
}

@article{agentgym,
  title={Agentgym: Evolving large language model-based agents across diverse environments},
  author={Xi, Zhiheng and Ding, Yiwen and Chen, Wenxiang and Hong, Boyang and Guo, Honglin and Wang, Junzhe and Yang, Dingwen and Liao, Chenyang and Guo, Xin and He, Wei and others},
  journal={arXiv preprint arXiv:2406.04151},
  year={2024}
}

@article{ppo,
  title={Proximal policy optimization algorithms},
  author={Schulman, John and Wolski, Filip and Dhariwal, Prafulla and Radford, Alec and Klimov, Oleg},
  journal={arXiv preprint arXiv:1707.06347},
  year={2017}
}

@inproceedings{rloo,
  title={Buy 4 reinforce samples, get a baseline for free!},
  author={Kool, Wouter and van Hoof, Herke and Welling, Max},
  booktitle={ICLR 2019 Workshop},
  year={2019},
}

@article{grpo,
  title={Deepseekmath: Pushing the limits of mathematical reasoning in open language models},
  author={Shao, Zhihong and Wang, Peiyi and Zhu, Qihao and Xu, Runxin and Song, Junxiao and Bi, Xiao and Zhang, Haowei and Zhang, Mingchuan and Li, YK and Wu, Yang and others},
  journal={arXiv preprint arXiv:2402.03300},
  year={2024}
}

@article{dr_grpo,
  title={Understanding R1-Zero-Like Training: A Critical Perspective},
  author={Liu, Zichen and Chen, Changyu and Li, Wenjun and Qi, Penghui and Pang, Tianyu and Du, Chao and Lee, Wee Sun and Lin, Min},
  journal={arXiv preprint arXiv:2503.20783},
  year={2025}
}

@article{istar,
  title={Agentic Reinforcement Learning with Implicit Step Rewards},
  author={Liu, Xiaoqian and Wang, Ke and Wu, Yuchuan and Huang, Fei and Li, Yongbin and Zhang, Junge and Jiao, Jianbin},
  journal={arXiv preprint arXiv:2509.19199},
  year={2025}
}

@article{deepseek-r1,
  title={Deepseek-r1: Incentivizing reasoning capability in llms via reinforcement learning},
  author={Guo, Daya and Yang, Dejian and Zhang, Haowei and Song, Junxiao and Zhang, Ruoyu and Xu, Runxin and Zhu, Qihao and Ma, Shirong and Wang, Peiyi and Bi, Xiao and others},
  journal={arXiv preprint arXiv:2501.12948},
  year={2025}
}

@article{spear,
  title={Learn the Ropes, Then Trust the Wins: Self-imitation with Progressive Exploration for Agentic Reinforcement Learning},
  author={Qin, Yulei and Tan, Xiaoyu and He, Zhengbao and Li, Gang and Lin, Haojia and Li, Zongyi and Xu, Zihan and Shi, Yuchen and Cai, Siqi and Rui, Renting and others},
  journal={arXiv preprint arXiv:2509.22601},
  year={2025}
}

@article{deepseek_v3,
  title={Deepseek-v3 technical report},
  author={Liu, Aixin and Feng, Bei and Xue, Bing and Wang, Bingxuan and Wu, Bochao and Lu, Chengda and Zhao, Chenggang and Deng, Chengqi and Zhang, Chenyu and Ruan, Chong and others},
  journal={arXiv preprint arXiv:2412.19437},
  year={2024}
}

@article{gpt-4o,
  title={Gpt-4o system card},
  author={Hurst, Aaron and Lerer, Adam and Goucher, Adam P and Perelman, Adam and Ramesh, Aditya and Clark, Aidan and Ostrow, AJ and Welihinda, Akila and Hayes, Alan and Radford, Alec and others},
  journal={arXiv preprint arXiv:2410.21276},
  year={2024}
}

@misc{gspo,
      title={Group Sequence Policy Optimization}, 
      author={Chujie Zheng and Shixuan Liu and Mingze Li and Xiong-Hui Chen and Bowen Yu and Chang Gao and Kai Dang and Yuqiong Liu and Rui Men and An Yang and Jingren Zhou and Junyang Lin},
      year={2025},
      eprint={2507.18071},
      archivePrefix={arXiv},
      primaryClass={cs.LG},
      url={https://arxiv.org/abs/2507.18071}, 
}

@misc{qwen25,
      title={Qwen2.5 Technical Report}, 
      author={An Yang and Baosong Yang and Beichen Zhang and Binyuan Hui and Bo Zheng and Bowen Yu and Chengyuan Li and Dayiheng Liu and Fei Huang and Haoran Wei and Huan Lin and Jian Yang and Jianhong Tu and Jianwei Zhang and Jianxin Yang and Jiaxi Yang and Jingren Zhou and Junyang Lin and Kai Dang and Keming Lu and Keqin Bao and Kexin Yang and Le Yu and Mei Li and Mingfeng Xue and Pei Zhang and Qin Zhu and Rui Men and Runji Lin and Tianhao Li and Tianyi Tang and Tingyu Xia and Xingzhang Ren and Xuancheng Ren and Yang Fan and Yang Su and Yichang Zhang and Yu Wan and Yuqiong Liu and Zeyu Cui and Zhenru Zhang and Zihan Qiu},
      year={2025},
      eprint={2412.15115},
      archivePrefix={arXiv},
      primaryClass={cs.CL},
      url={https://arxiv.org/abs/2412.15115}, 
}

@misc{qwen3,
      title={Qwen3 Technical Report}, 
      author={An Yang and Anfeng Li and Baosong Yang and Beichen Zhang and Binyuan Hui and Bo Zheng and Bowen Yu and Chang Gao and Chengen Huang and Chenxu Lv and Chujie Zheng and Dayiheng Liu and Fan Zhou and Fei Huang and Feng Hu and Hao Ge and Haoran Wei and Huan Lin and Jialong Tang and Jian Yang and Jianhong Tu and Jianwei Zhang and Jianxin Yang and Jiaxi Yang and Jing Zhou and Jingren Zhou and Junyang Lin and Kai Dang and Keqin Bao and Kexin Yang and Le Yu and Lianghao Deng and Mei Li and Mingfeng Xue and Mingze Li and Pei Zhang and Peng Wang and Qin Zhu and Rui Men and Ruize Gao and Shixuan Liu and Shuang Luo and Tianhao Li and Tianyi Tang and Wenbiao Yin and Xingzhang Ren and Xinyu Wang and Xinyu Zhang and Xuancheng Ren and Yang Fan and Yang Su and Yichang Zhang and Yinger Zhang and Yu Wan and Yuqiong Liu and Zekun Wang and Zeyu Cui and Zhenru Zhang and Zhipeng Zhou and Zihan Qiu},
      year={2025},
      eprint={2505.09388},
      archivePrefix={arXiv},
      primaryClass={cs.CL},
      url={https://arxiv.org/abs/2505.09388}, 
}

@misc{tuluv3,
      title={Tulu 3: Pushing Frontiers in Open Language Model Post-Training}, 
      author={Nathan Lambert and Jacob Morrison and Valentina Pyatkin and Shengyi Huang and Hamish Ivison and Faeze Brahman and Lester James V. Miranda and Alisa Liu and Nouha Dziri and Shane Lyu and Yuling Gu and Saumya Malik and Victoria Graf and Jena D. Hwang and Jiangjiang Yang and Ronan Le Bras and Oyvind Tafjord and Chris Wilhelm and Luca Soldaini and Noah A. Smith and Yizhong Wang and Pradeep Dasigi and Hannaneh Hajishirzi},
      year={2025},
      eprint={2411.15124},
      archivePrefix={arXiv},
      primaryClass={cs.CL},
      url={https://arxiv.org/abs/2411.15124}, 
}

@inproceedings{webshop,
	title = {{WebShop}: {Towards} {Scalable} {Real}-{World} {Web} {Interaction} with {Grounded} {Language} {Agents}},
	volume = {35},
	url = {https://proceedings.neurips.cc/paper_files/paper/2022/file/82ad13ec01f9fe44c01cb91814fd7b8c-Paper-Conference.pdf},
	booktitle = {Advances in {Neural} {Information} {Processing} {Systems}},
	publisher = {Curran Associates, Inc.},
	author = {Yao, Shunyu and Chen, Howard and Yang, John and Narasimhan, Karthik},
	editor = {Koyejo, S. and Mohamed, S. and Agarwal, A. and Belgrave, D. and Cho, K. and Oh, A.},
	year = {2022},
	pages = {20744--20757},
}

@inproceedings{ttpa,
    title = "{TTPA}: Token-level Tool-use Preference Alignment Training Framework with Fine-grained Evaluation",
    author = "Huang, Chengrui  and
      Gao, Shen  and
      Shi, Zhengliang  and
      Wang, Dongsheng  and
      Shang, Shuo",
    editor = "Christodoulopoulos, Christos  and
      Chakraborty, Tanmoy  and
      Rose, Carolyn  and
      Peng, Violet",
    booktitle = "Findings of the Association for Computational Linguistics: EMNLP 2025",
    month = nov,
    year = "2025",
    address = "Suzhou, China",
    publisher = "Association for Computational Linguistics",
    url = "https://aclanthology.org/2025.findings-emnlp.882/",
    doi = "10.18653/v1/2025.findings-emnlp.882",
    pages = "16240--16255",
    ISBN = "979-8-89176-335-7"
}

@inproceedings{toolleanringinthewild,
author = {Shi, Zhengliang and Gao, Shen and Yan, Lingyong and Feng, Yue and Chen, Xiuyi and Chen, Zhumin and Yin, Dawei and Verberne, Suzan and Ren, Zhaochun},
title = {Tool Learning in the Wild: Empowering Language Models as Automatic Tool Agents},
year = {2025},
isbn = {9798400712746},
publisher = {Association for Computing Machinery},
address = {New York, NY, USA},
url = {https://doi.org/10.1145/3696410.3714825},
doi = {10.1145/3696410.3714825},
booktitle = {Proceedings of the ACM on Web Conference 2025},
pages = {2222–2237},
numpages = {16},
location = {Sydney NSW, Australia},
series = {WWW '25}
}

@inproceedings{cooperativeandinteractiveagents,
    title = "Learning to Use Tools via Cooperative and Interactive Agents",
    author = "Shi, Zhengliang  and
      Gao, Shen  and
      Chen, Xiuyi  and
      Feng, Yue  and
      Yan, Lingyong  and
      Shi, Haibo  and
      Yin, Dawei  and
      Ren, Pengjie  and
      Verberne, Suzan  and
      Ren, Zhaochun",
    editor = "Al-Onaizan, Yaser  and
      Bansal, Mohit  and
      Chen, Yun-Nung",
    booktitle = "Findings of the Association for Computational Linguistics: EMNLP 2024",
    month = nov,
    year = "2024",
    address = "Miami, Florida, USA",
    publisher = "Association for Computational Linguistics",
    url = "https://aclanthology.org/2024.findings-emnlp.624/",
    doi = "10.18653/v1/2024.findings-emnlp.624",
    pages = "10642--10657"
}

@inproceedings{Confucius,
author = {Gao, Shen and Shi, Zhengliang and Zhu, Minghang and Fang, Bowen and Xin, Xin and Ren, Pengjie and Chen, Zhumin and Ma, Jun and Ren, Zhaochun},
title = {Confucius: iterative tool learning from introspection feedback by easy-to-difficult curriculum},
year = {2024},
isbn = {978-1-57735-887-9},
publisher = {AAAI Press},
url = {https://doi.org/10.1609/aaai.v38i16.29759},
doi = {10.1609/aaai.v38i16.29759},
booktitle = {Proceedings of the Thirty-Eighth AAAI Conference on Artificial Intelligence and Thirty-Sixth Conference on Innovative Applications of Artificial Intelligence and Fourteenth Symposium on Educational Advances in Artificial Intelligence},
articleno = {2011},
numpages = {9},
series = {AAAI'24/IAAI'24/EAAI'24}
}

@misc{infiguir1,
      title={InfiGUI-R1: Advancing Multimodal GUI Agents from Reactive Actors to Deliberative Reasoners}, 
      author={Yuhang Liu and Pengxiang Li and Congkai Xie and Xavier Hu and Xiaotian Han and Shengyu Zhang and Hongxia Yang and Fei Wu},
      year={2025},
      eprint={2504.14239},
      archivePrefix={arXiv},
      primaryClass={cs.AI},
      url={https://arxiv.org/abs/2504.14239}, 
}

@inproceedings{cot,
	title = {Chain-of-{Thought} {Prompting} {Elicits} {Reasoning} in {Large} {Language} {Models}},
	volume = {35},
	url = {https://proceedings.neurips.cc/paper_files/paper/2022/file/9d5609613524ecf4f15af0f7b31abca4-Paper-Conference.pdf},
	booktitle = {Advances in {Neural} {Information} {Processing} {Systems}},
	publisher = {Curran Associates, Inc.},
	author = {Wei, Jason and Wang, Xuezhi and Schuurmans, Dale and Bosma, Maarten and ichter, brian and Xia, Fei and Chi, Ed and Le, Quoc V and Zhou, Denny},
	editor = {Koyejo, S. and Mohamed, S. and Agarwal, A. and Belgrave, D. and Cho, K. and Oh, A.},
	year = {2022},
	pages = {24824--24837},
}
\clearpage
\appendix
\section{Supplementary Results of Experiments}\label{app:additional_detail_exp}
\begin{table*}[t]
\centering
\resizebox{\textwidth}{!}{
\begin{tabular}{lccccccccc|cccc}
\toprule
\multirow{2}{*}{Method} & \multicolumn{9}{c}{\textbf{ALFWorld}} & \multicolumn{4}{c}{\textbf{ScienceWorld}} \\  
& Pick & Cool & Heat & Look & Pick2 & Clean & All & OOD & Avg. & L0 & L1 & L2 & Avg. \\
\midrule
\multicolumn{13}{l}{\textit{Qwen2.5-7B-Instruct}} \\
~ColdStart & \textbf{100.0} &84.0 & \textbf{100.0} &\underline{92.3} &\underline{91.7} &92.6 &93.6 &\underline{97.8} &95.7 &\underline{66.5} &62.2 &48.1 &58.9 \\
~\model &\textbf{100.0} &\textbf{100.0} &\underline{93.8} & \textbf{100.0} & \textbf{100.0} &\underline{96.3} &\textbf{98.6} &\underline{97.8} &\textbf{98.2} &\textbf{66.6} &\textbf{66.5} &\textbf{51.0} &\textbf{61.4} \\
~~w/o DAR & \textbf{100.0} & \underline{96.0} & \underline{93.8} & 84.6 & \textbf{100.0} & \textbf{100.0} & \underline{97.1} & 97.0 & 97.1 & 65.1 & 62.8 & 49.0 & 59.0 \\
~~w/o DTR & \textbf{100.0} & \textbf{100.0} & \textbf{100.0} & \underline{92.3} & 87.5 & \underline{96.3} & 96.4 & \textbf{98.5} & \underline{97.5} & 64.4 & 63.6 & 49.7 & 59.2 \\
~~w/o DAR \& DTR &\underline{97.1} &92.0 & \textbf{100.0} & \textbf{100.0} &\underline{91.7} & \textbf{100.0} &96.4 &\textbf{98.5} &\underline{97.5} &66.3 &\underline{65.7} &\underline{49.9} &\underline{60.6} \\
\bottomrule

\end{tabular}
}
\caption{Detailed per-task success rate of \model variants on ALFWorld In Domain split, alongside overall results on ScienceWorld. This table complements the summary success rate in Table~\ref{tab:ablation_study} by providing fine-grained performance across individual ALFWorld tasks. \textbf{Bold scores} indicate the best performance, while \underline{underlined scores} represent the second-best.}
\label{tab:detail_ablation_study}
\end{table*}

\subsection{Additional Results of Scaling Analysis}\label{app:scaling_qwen3}

\begin{figure}[ht] 
\centering
  \includegraphics[width=\linewidth]{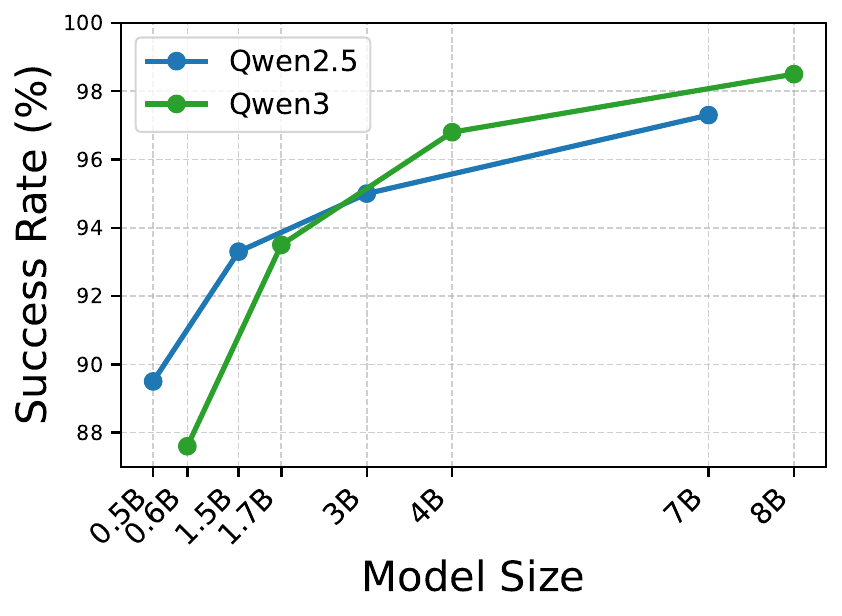} 
  \caption{Scaling experiments conducted on Qwen3.}
  \label{fig:qwen3_scaling}
\end{figure}

In Section~\ref{sec:scaling_abilities}, we presented the performance of Qwen2.5 variants across different model scales. Here, we provide a detailed results of Qwen3 variants with varying parameter, reporting their average success rate on both seen and unseen tasks of ALFWorld. Full results are shown in Figure~\ref{fig:qwen3_scaling}.
As illustrated, stronger models consistently achieve higher performance. Within the Qwen3 series, success rate improves steadily with model size, demonstrating clear and predictable scaling behavior.

\subsection{Detailed Results of Ablation Study}\label{app:ablation_details}
In Section~\ref{sec:ablation_study}, we have reported the performance of various \model variants. In this section, we provide a more detailed breakdown of their results across individual tasks in the ALFWorld In-Domain setting. Full scores are shown in Table~\ref{tab:detail_ablation_study}.

\subsection{Details of Training Efficiency Analysis}~\label{app:time_details} 
Both our approach and GiGPO are adapted from GRPO and thus follow the same training pipeline as GRPO, which includes multi-turn rollout, old policy probability computation, reward and advantage estimation, and policy updates. We have reported the total training time in Section~\ref{sec:train_efficiency}; in this section, we detail the time consumption of different training components per step.

During \model training, the agent is allowed to explore up to 4 parallel environments per step, with a maximum of 25 steps per trajectory. In contrast, GiGPO explores only a single environment with no more than 50 steps. Under this setting, Figure~\ref{fig:time_details} indicates that although the \model rollout phase takes longer than that of GiGPO due to parallel execution, policy updates, probability recomputation, and advantage estimation are consistently faster. We attribute this to the reduced average trajectory depth under parallel exploration, which yields fewer but higher-quality samples for policy updates.

\subsection{Reward Hacking Analysis}
\begin{table*}[t]
\centering
\resizebox{0.95\textwidth}{!}{
\begin{tabular}{ccccc}
\toprule
Train Step & Number of Action (Mean) & Number of Action (std) & Token(Response) & Success Rate \\
\midrule
50 & 2.2513 & 1.9059 & 2245.0 & 87.86 \\
100 & 2.1968 & 1.7134 & 2199.0 & 91.43 \\
200 & 2.1817 & 1.7817 & 2188.6 & 93.57 \\
300 & 2.3551 & 1.7609 & 2013.0 & 91.43 \\
400 & 2.1753 & 1.6628 & 2153.3 & 95.00 \\
500 & 2.1827 & 1.5964 & 2193.4 & 95.71 \\
\bottomrule
\end{tabular}
} 
\caption{Training Dynamics of DPEPO}
\label{tab:reward_hacking}
\end{table*}

Although we do not explicitly penalize over-exploration in the reward design, the inherent context window imposes a practical constraint. Excessive exploration across too many environments causes a rapid increase in context length, and exceeding the context limit is treated as task failure. Additionally, exploring too many environments leads to redundant actions and state transitions. In this situation, the diversity-driven step-level reward naturally penalizes this behavior, which help reduce unnecessary exploration.
We further present the training dynamics of Qwen2.5-1.5B-Instruct to substantiate above claim. As shown in Table \ref{tab:reward_hacking}, the agent's exploration behavior remains stable.

\subsection{DPEPO's Performance without ColdStart}
\begin{table*}[t]
\centering
\resizebox{0.95\textwidth}{!}{
\begin{tabular}{lccccccccc}
\toprule
\textbf{Algorithm} & \textbf{Pick} & \textbf{Cook} & \textbf{Heat} & \textbf{Look} & \textbf{Pick2} & \textbf{Clean} & \textbf{All} & \textbf{OOD} & \textbf{Avg} \\
\midrule
GRPO            & 90.8   & 72.5   & 74.7   & 66.1   & 77.6   & 89.3   & 77.6   & 77.3   & 77.5 \\
DPEPO w/o ColdStart & 100.0  & 92.0   & 87.5   & 100.0  & 66.667 & 85.185 & 88.571 & 91.8   & 90.2 \\
DPEPO           & 100.0  & 100.0  & 93.8   & 100.0  & 100.0  & 96.3   & 98.6   & 97.8   & 98.2 \\
\bottomrule
\end{tabular}
} 
\caption{Performance Comparison of DPEPO, DPEPO without ColdStart (DPEPO w/o ColdStart), and GRPO.}
\label{tab:dpepo_wo_coldstart}
\end{table*}
A concern may arise regarding the respective contributions of the SFT and RL stages. Therefore, we supply additional ablation results conducted on Qwen2.5-7B-Instruct under the same setting as GRPO. As shown in Table \ref{tab:dpepo_wo_coldstart}, the variant DPEPO w/o ColdStart still outperforms the GRPO baseline, confirming the effectiveness of DPEPO.

\subsection{Comparision with Stronger Baselines}
\begin{table}[t]
\centering
\resizebox{\columnwidth}{!}{
\begin{tabular}{lccccccccc}
\toprule
\textbf{Method} & \textbf{In Domain} & \textbf{OOD} & \textbf{Avg} \\
\midrule
SFT & 43.0 & 38.7 & 40.85 \\
AgentGYM (SFT + DPO) & 76.6 & 63.3 & 69.95 \\
ETO (SFT + DPO) & 70.3 & 74.2 & 72.25 \\
MCTS & 56.4 & 55.2 & 55.80 \\
MCTS + SFT & 91.4 & 92.5 & 91.95 \\
MCTS + DPO & 91.4 & 93.3 & 92.35 \\
DPEPO & 98.6 & 97.8 & 98.20 \\
\bottomrule
\end{tabular}
} 
\caption{Comparison with stronger baselines, the experiments were conducted on Qwen2.5-7B-Instruct.}
\label{tab:compare_w_mcts}
\end{table}
As presented in Table \ref{tab:compare_w_mcts}, we further added performance comparisons on ALFWorld against stronger baselines, including methods that employ exploration-based strategies and advanced imitation learning approaches. Although basic SFT yields only moderate performance, prompting with advanced exploration strategies like MCTS brings substantial improvements. Furthermore, curating trajectories or preference pairs from MCTS traces to fine-tune the LLM enhances performance even further. Despite these advancements, these methods still lag behind DPEPO and suffer from complex backtracking and expansion operations.

\subsection{Parallel Exploration Behavior Analysis}
\begin{figure}[ht] 
\centering
  \includegraphics[width=\linewidth]{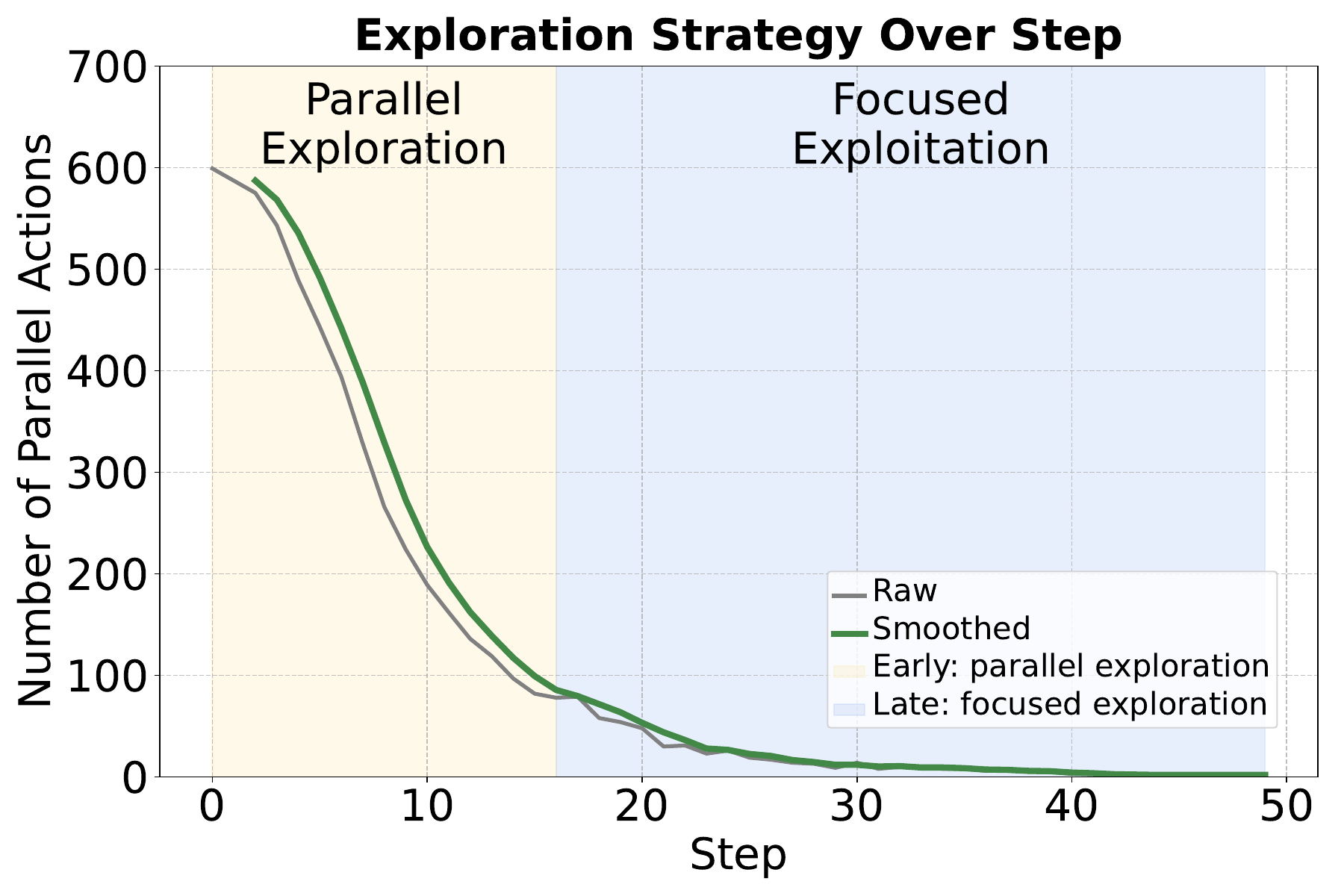} 
  \caption{Agent's Exploration Strategy Dynamics}
  \label{fig:exploration_case}
\end{figure}
\paragraph{Exploration Strategy} 
As presented in Figure~\ref{fig:exploration_case}, the agent performs more parallel actions in the early steps. At this stage, the agent is unfamiliar with the environment and relies on parallel exploration to rapidly build a preliminary understanding of task and environment.
In the later steps, after acquiring a basic model of the environment and the task, the agent shifts its strategy toward focused exploration. It reduces the number of parallel actions and concentrates on a specific environment, using only a few parallel actions to support its more targeted exploration.

\paragraph{Efficient Search via Parallel Actions} 
The case shown in Figure~\ref{fig:parallel_explore} illustrates how the agent efficiently searches for a candle that is not visible in the initial observation. Recognizing that candles are commonly stored on shelves, the agent prioritizes the three shelves in the room as likely candidates. Instead of checking them one by one, it leverages parallel environments to visit shelf 1, shelf 2, and shelf 3 simultaneously. This strategy reduces search time and demonstrates the agent's ability to perform goal-directed exploration by distributing candidate actions across multiple environments. 

\paragraph{Validating Hypotheses via Parallel Exploration} As the case shown in Figure~\ref{fig:validate_via_paralllel}, when the agent repeatedly fails to take book 2 from the bed, it hypothesizes that there might be a carrying limit. To test this, it uses parallel environments to check its inventory in env\_1 and env\_2 while also looking around in env\_3 to gather more context. This demonstrates the model’s ability to perform reasoning and efficiently validate hypotheses through parallel exploration.

\paragraph{Keep Exploring Even After Finding a Solution}
The case in Figure~\ref{fig:keep_explore} shows that even after finding the mug in environment 4, the agent keeps checking other places in parallel, just in case there’s a better option or something it missed. This shows the model doesn’t stop exploring too early, which helps it make better decisions in uncertain environments.

\paragraph{Structured Search Across Location Types}
The case in Figure~\ref{fig:structure_explore} shows that after failing to find a tomato in the fridge across all environments, the agent, based on common-sense knowledge of where tomatoes might be,  systematically uses each parallel environment to explore a different type of location: cabinets, drawers, and countertops. Instead of randomly trying places, it assigns each environment a distinct category to cover more ground efficiently. This demonstrates the \model's ability to perform structured exploration.

\subsection{Applicability of DPEPO Analysis}
Although parallelizing interactions remains difficult in embodied agent scenarios, our paradigm is highly applicable to a wide range of dominant current agent scenarios.
(1) Code Generation: Agents can compile and test code in isolated, parallel sanboxes to rapidly verify correctness.
(2) Tool Learning: Agents mastering multiple tools and APIs can invoke them in parallel to support simultaneous sub-task execution or efficient trial-and-error.
(3) GUI Agent: Web or desktop environments can be instantiated simultaneously in multiple headless browsers or virtual machines.
(4) DeepResearch Agent: Information seeking tasks naturally support parallel querying across diverse search engines and document sources. We hope these examples illustrate that, though challenges remain in real-world embodied scenarios, our approach still offers a practical and efficient solution for many important agent tasks. We look forward to further exploring how this paradigm might be adapted to broader scenarios in future work.

\section{Experimental Details} \label{app:experiment_details}
\subsection{SFT Training Details}
\paragraph{Datasets}
For ALFWorld, we annotate 500 successful trajectories from the training split.
For ScienceWorld, we adopt the same data partitioning strategy as RLVMR, which defines three difficulty levels: L0, L1, and L2. For training on L0 and L1, we annotate 1,000 trajectories from the L1 training set (used for both L0 and L1 training for simplicity). Consequently, our model faces a more challenging L0 evaluation compared to RLVMR, as it is trained with fewer L0-specific examples. For L2, we annotate 500 trajectories for training.

\paragraph{Experiment and Hyperparameters Details}
All our SFT experiments are conducted using the OpenRLHF framework. Specifically, we employ DeepSpeed ZeRO-3 with Adam optimizer offloading enabled. The model is trained with a maximum sequence length of 8,192 tokens, a learning rate of 5e‑6, and bfloat16 mixed-precision, with FlashAttention-2 enabled in all experiments. The batch size is set to 4 for ALFWorld and 16 for ScienceWorld.

\subsection{RL Training Details}
\paragraph{Datasets} 
For both ALFWorld and ScienceWorld, we select 500 samples from the training set that excludes SFT samples for RL training.

\paragraph{Experiment and Hyperparameters Details}
All RL experiments is conducted based on verl-agent framework and Ray-based dirtributed parallel environment. The maximum prompt length is set to 10,240 tokens and the maximum response length to 2,048 tokens, with each trajectory allowed up to 25 environment steps. The final actor learning rate is 1e-6, and we employ parallel trajectory-level reward that awards +1 for task success. Repetition penalties are applied consistently across both environments, with coefficients of 0.8 for depth-level action repetition and 0.95 for width-level action repetition, depth-level transition repetition, and width-level transition repetition. Notably, ScienceWorld includes an additional penalty coefficient of 0.5 for invalid actions, which is not used in ALFWorld. Both environments run with a maximum of 4 parallel exploration trajectories during training. For group-based reinforcement learning, ALFWorld adopts a group size of 8 while ScienceWorld uses a group size of 4; in both cases, 4 distinct groups are sampled per rollout. The rollout temperature is set to 0.4, and the mini-batch size is 32. All policy updates are performed using LoRA, with the LoRA rank and scaling factor are all set to 32.

\subsection{Computation Details.}
For parallel environments, we cap the maximum CPU usage at 0.3.  
For training on ALFWorld and ScienceWorld, experiments with Qwen2.5-1.5B are conducted on 4×L40 GPUs, and those with Qwen2.5-7B on 4×H20 GPUs. Each model is trained for 1 epoch (500 samples) with 125 iterations.

\section{Prompt}\label{app:prompts}
\subsection{Prompt Details for Training}
\paragraph{System Prompt}
System prompt is crucial for prompt engineering. The LLM follows the instructions and assumes the role specified in the it, such as translator, programmer, etc. Therefore, as illustrated in Figure~\ref{prompt:system_prompt}, we have provided a detailed description of the desired behaviors of the parallel agent within the system prompt to facilitate its parallel exploration and thereby accomplish the task better.
Specifically, similar to ReAct, parallel agent is required to engage in reasoning before making a decision. For example, it needs to consider which actions can be executed in parallel to gather more information and which environments warrant further exploration. This reasoning process must be enclosed within \texttt{the <think></think>} tags. After reasoning, the agent can take actions. Unlike ReAct, the parallel agent can interact with multiple environments simultaneously. Thus, we require it to specify environment-specific actions within \texttt{<parallel></parallel>} tags, with each individual action wrapped in \texttt{<env\_i><env>} tags to clearly indicate which environment the action belongs to and to enable easier parsing of the actions. 
\paragraph{Prompt for First Step} 
At the beginning of a task, the agent has not yet interacted with the environment. The environment provides only initial state information, a set of available actions, and the task description. We organize this information into a concise yet clear prompt (As shown in Figure~\ref{prompt:prompt_first_step}) to help the agent understand the initial context. Additionally, for ALF-World, we do not include explanations for the available actions. However, for the ScienceWorld dataset, which involves combinations of actions and objects, we introduce brief action descriptions to avoid excessive token consumption.

\paragraph{Prompt for Intermediate Step} 
As the number of parallel agent exploration steps increases, maintaining the full history of interactions becomes costly in terms of both memory and computation. To address this, at step $t$, for information from steps $1$ to $t-2$, we retain only the actions and their corresponding observations within \{history\_info\}. In contrast, for the immediate previous step $t-1$, we preserve all of the parallel actions, observations, and executable actions within \{last\_history\}. 
The details of \{history\_info\} and \{last\_history\} are presented in Figure~\ref{prompt:history_info} and Figure~\ref{prompt:last_history}.
With this carefully designed prompt presented in Figure~\ref{prompt:intermediate_step}, the context length is reduced while still retaining sufficient information for the agent to perform its task more effectively.

\paragraph{Exploration-limiting prompt}
For experiments that require controlling the number of parallel environments, we designed a prompt to limit the maximum number of environments the model can explore. In this prompt, we specify the allowed number of environments using ``env\_num'' and inform the model that if it attempts to explore more than this number, the entire task will be considered a failure.
In addition to this prompt-based restriction, we also enforce behavioral constraints: if the model tries to access an environment that is not permitted, the entire task is immediately seen as failed.

\paragraph{Action-Explanation Prompt}
Since the tasks in ScienceWorld require actions to interact with objects and involves a large number of both actions and objects, it is impractical to include all possible object-action combinations in the limited context. Therefore, we adopt the same prompt as RLVMR, which provides only a list of executable actions along with explanations of their specific details and the objects they can interact with. This prompt is illustrated in Figure~\ref{prompt:action_explaination}.

\begin{figure*}
  \centering
  \includegraphics[width=\linewidth]{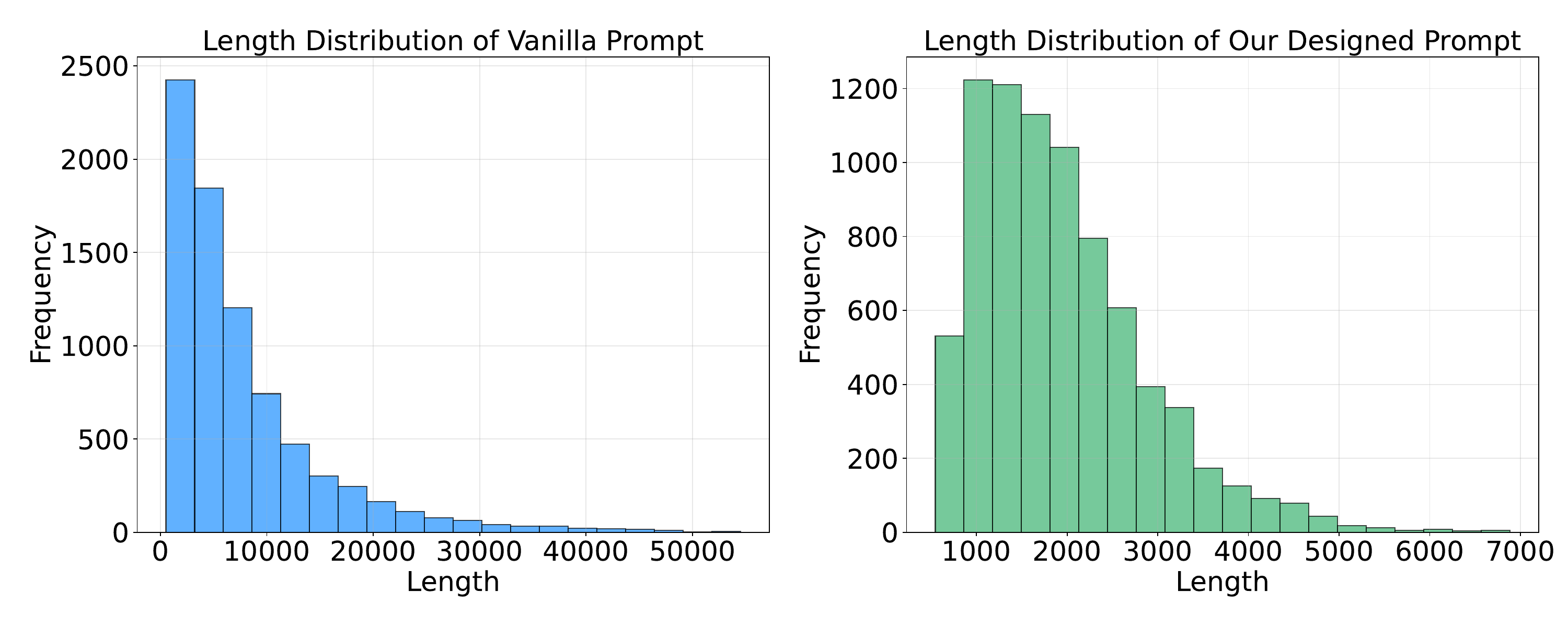} 
  \caption{Context length distribution for the vanilla prompt (left) and our designed prompt (right).}
  \label{fig:length_distribution}
\end{figure*}

\subsection{Effectiveness of Designed Prompts}
\paragraph{Token Budget Analysis}
When tackling long-horizon tasks, traditional methods typically append all historical information and intermediate reasoning steps into a single context window. However, as the number of steps increases, the context continuously expands. In this section, we evaluate the effectiveness of our designed prompt in compressing the context window. As illustrated in Figure~\ref{fig:length_distribution}, our method can compress prompts that originally contain at most 50,000 tokens down to at most 7,000 tokens, significantly reducing the token budget while still maintaining the agent’s strong performance.

\paragraph{Comparison of Context Management Strategies}
GiGPO employs a similar prompting strategy, but instead of retaining all historical information, it only preserves the most recent few steps within a limited context window. 
We evaluate GiGPO's performance with varying context window sizes, as shown in Table~\ref{tab:varing_size_gigpo}. Although the context length may be shorter, its performance does not improve as the context window expands. Furthermore, once the number of retained steps exceeds 16, the total context length will surpasses the model’s maximum sequence length, leading to truncation.

In contrast, our proposed prompting design not only preserves more comprehensive historical information but also adheres to the token budget. Furthermore, as demonstrated in Section~\ref{sec:scaling_abilities} and Section~\ref{sec:explore_efficient_analysis}, under our prompting scheme, the performance of our method consistently improves as the number of environments increases, highlighting its efficiency, robustness and scalability.

\begin{figure*}[h] 
\begin{tcolorbox}[
    colback=gray!5,
    colframe=gray!150,
    coltitle=white,
    title=System Prompt,
    fonttitle=\bfseries,
    boxrule=1.5pt,
    top=4pt, bottom=4pt, left=8pt, right=8pt,
    width=\textwidth, 
]
You are an expert agent operating in the \{ALFRED / ScienceWorld\} embodied environment.
Given a task, you need to reason first in your mind.
Your reasoning process must be enclosed within <think> </think> tags,
for example: <think> reasoning process here </think>.

After thinking, you may take actions. You can either explore multiple parallel environments with multiple actions or take an action in a specific environment.
At the very beginning, every environment have the same status, but each environment is independent, they do not share state changes after actions are taken.
So, parallel actions are executed simultaneously across different environments. The parallel actions are not carried out sequentially.
You must wrap each action in specific environment tags like 
\texttt{<env\_i>} \ldots{} \texttt{</env\_i>} 
to indicate which environment you are acting in.

To take multiple actions at the same time in different environments, 
use the \texttt{<parallel>} \ldots{} \texttt{</parallel>} tags and wrap 
each action within its corresponding \texttt{<env\_i>} \ldots{} 
\texttt{</env\_i>} tag, where \(i\) refers to the \(i\)-th environment:

\begin{verbatim}
<parallel>
<env_1> possible action 1 </env_1>
...
<env_i> possible action 2 </env_i>
</parallel>
\end{verbatim}

Your output must follow the rules above.

\end{tcolorbox}
\caption{The system prompt template.}
\label{prompt:system_prompt}
\end{figure*}
\begin{figure*}[t] 
\begin{tcolorbox}[
    colback=gray!5,
    colframe=gray!150,
    coltitle=white,
    title=Prompt for First Step,
    fonttitle=\bfseries,
    boxrule=1.5pt,
    top=4pt, bottom=4pt, left=8pt, right=8pt,
    width=\textwidth, 
]
You are an expert agent operating in the \{ALFRED / ScienceWorld\} Embodied Environment.  

Your task is to: \texttt{\{task\_description\}}  

Your current observation is: \texttt{\{current\_observation\}}  

Your admissible actions in the current situation are: \texttt{\{admissible\_actions\}}.

Your output must follow the rules above.
\end{tcolorbox}
\caption{The prompt template at the beginning of a task.}
\label{prompt:prompt_first_step}
\end{figure*}

\begin{figure*}[t] 
\begin{tcolorbox}[
    colback=gray!5,
    colframe=gray!150,
    coltitle=white,
    title=Prompt for Intermediate Step,
    fonttitle=\bfseries,
    boxrule=1.5pt,
    top=4pt, bottom=4pt, left=8pt, right=8pt,
    width=\textwidth, 
]

You are an expert agent operating in the \{ALFRED / ScienceWorld\} Embodied Environment.  

Your task is to: \texttt{\{task\_description\}}.  

Your initial observation is: \texttt{\{initial\_observation\}}.

\texttt{\{history\_info\}}.

In your last step, your actions, corresponding observations, and admissible actions are:  
\texttt{\{last\_history\}}
\end{tcolorbox}
\caption{Careful designed prompt for intermediate step.}
\label{prompt:intermediate_step}
\end{figure*}

\begin{figure*}[t] 
\begin{tcolorbox}[
    colback=gray!5,
    colframe=gray!150,
    coltitle=white,
    title=history\_info,
    fonttitle=\bfseries,
    boxrule=1.5pt,
    top=4pt, bottom=4pt, left=8pt, right=8pt,
    width=\textwidth, 
]
You have already taken multiple actions in multiple parallel environments. Below are the most recent observations and the corresponding actions you took:

In Environment 1\\
Action 1: \{action\_1\}\\
Observation 1: \{obs\_1\}\\
Action 2: \{action\_2\}\\
Observation 2: \{obs\_2\}\\

In Environment 2\\
Action 1: \{action\_1\}\\
Observation 1: \{obs\_1\}\\
Action 2: \{action\_2\}\\
Observation 2: \{obs\_2\}\\
Action 3: \{action\_3\}\\
Observation 3: \{obs\_3\}

\end{tcolorbox}
\caption{Historical information used in the ``Prompt for Intermediate Step'' demonstration.}
\label{prompt:history_info}
\end{figure*}

\begin{figure*}[t] 
\begin{tcolorbox}[
    colback=gray!5,
    colframe=gray!150,
    coltitle=white,
    title=last\_history,
    fonttitle=\bfseries,
    boxrule=1.5pt,
    top=4pt, bottom=4pt, left=8pt, right=8pt,
    width=\textwidth, 
]
In Environment 1\\
Action 1: \{action\_1\}\\
Observation 1: \{obs\_1\}\\
Action 2: \{action\_2\}\\
Observation 2: \{obs\_2\}\\
Next Possible Actions: \{poa\_2\}\\
\\
In Environment 2\\
Action 1: \{action\_1\}\\
Observation 1: \{obs\_1\}\\
Next Possible Actions: \{poa\_1\}

\end{tcolorbox}
\caption{Last step information used in the ``Prompt for Intermediate Step'' demonstration.}
\label{prompt:last_history}
\end{figure*}

\begin{figure*}[t] 
\begin{tcolorbox}[
    colback=gray!5,
    colframe=gray!150,
    coltitle=white,
    title=Exploration\_limiting Prompt ,
    fonttitle=\bfseries,
    boxrule=1.5pt,
    top=4pt, bottom=4pt, left=8pt, right=8pt,
    width=\textwidth, 
]
You can explore up to \{env\_num\} different environments, ranging from 1 to \{env\_num\}. If you explore more than \{env\_num\} parallel environments, the task seens failed.

\end{tcolorbox}
\caption{Prompt for limiting the model to exploring at most \texttt{\{env\_num\}} environments.}
\label{prompt:explore_limti_prompt}
\end{figure*}

\begin{figure*}[t] 
\begin{tcolorbox}[
    colback=gray!5,
    colframe=gray!150,
    coltitle=white,
    title=Action Explanation Prompt for ScienceWorld ,
    fonttitle=\bfseries,
    boxrule=1.5pt,
    top=4pt, bottom=4pt, left=8pt, right=8pt,
    width=\textwidth, 
]
Here the explanations of actions you may take: \\
\{\{"action": "open OBJ", "description": "open a container"\}\}, \\
\{\{"action": "close OBJ", "description": "close a container"\}\}, \\
\{\{"action": "activate OBJ", "description": "activate a device"\}\}, \\
\{\{"action": "deactivate OBJ", "description": "deactivate a device"\}\}, \\
\{\{"action": "connect OBJ to OBJ", "description": "connect electrical components"\}\}, \\
\{\{"action": "disconnect OBJ", "description": "disconnect electrical components"\}\}, \\
\{\{"action": "use OBJ", "description": "use a device/item"\}\}, \\
\{\{"action": "look around", "description": "describe the current room"\}\}, \\
\{\{"action": "look at OBJ", "description": "describe an object in detail"\}\}, \\
\{\{"action": "look in OBJ", "description": "describe a container's contents"\}\}, \\
\{\{"action": "read OBJ", "description": "read a note or book"\}\}, \\
\{\{"action": "move OBJ to OBJ", "description": "move an object to a container"\}\}, \\
\{\{"action": "pick up OBJ", "description": "move an object to the inventory"\}\}, \\
\{\{"action": "put down OBJ", "description": "drop an inventory item"\}\}, \\
\{\{"action": "pour OBJ into OBJ", "description": "pour a liquid into a container"\}\}, \\
\{\{"action": "dunk OBJ into OBJ", "description": "dunk a container into a liquid"\}\},  \\
\{\{"action": "mix OBJ", "description": "chemically mix a container"\}\}, \\
\{\{"action": "go to LOC", "description": "move to a new location"\}\}, \\
\{\{"action": "eat OBJ", "description": "eat a food"\}\}, \\
\{\{"action": "flush OBJ", "description": "flush a toilet"\}\}, \\
\{\{"action": "focus on OBJ", "description": "signal intent on a task object"\}\}, \\
\{\{"action": "wait", "description": "take no action for 10 iterations"\}\}, \\
\{\{"action": "wait1", "description": "take no action for 1 iteration"\}\}, \\
\{\{"action": "task", "description": "describe current task"\}\}, \\
\{\{"action": "inventory", "description": "list your inventory"\}\}
\end{tcolorbox}
\caption{Action explanation prompt used in ScienceWorld, listing all valid environment interaction commands and their natural-language descriptions.}
\label{prompt:action_explaination}
\end{figure*}

\section{Pseudo Code for \model}\label{sec:pseudo_code}
To clearly illustrate the overall pipeline of our proposed \model, this section provides the detailed algorithmic pseudo code, as shown in Table~\ref{tab:pseudo_code}.

\begin{algorithm*}[t]
\caption{Training LLM Agents with \model}
\label{alg:gigpo}
\begin{algorithmic}[1]
\State {\bfseries Require:} Initial policy $\pi_{\theta_{\text{old}}}$, task distribution $p(X)$, group size $N$, number of parallel environment $K$
\For{each training iteration}
    \State Update the old policy model: $\theta_{\text{old}} \leftarrow \theta$
    \State \small{\color{gray}{// Multi-turn parallel rollout stage}}
    \State Sample task $x \sim p(X)$ and initialize $N$ parallel environment workers, each worker manages $K$ identical environments.
    \For{$t = 1$ to $T$}
        \State Sample parallel actions $\bigl\{A_{i,t} \sim \pi_{\theta_{\text{old}}}\bigr\}_{i=1}^N$
        \State Execute parallel actions , observe parallel observations $\{S_{i,t+1}\}_{i=1}^N$
    \EndFor
    
    \State \small{\color{gray}{// Advantage computation stage}}
    \State Compute trajectory relative advantages $\Phi_{traj}(\tau_i)$ via Equation~(\ref{eq:trajectory_advantage})
    \State Compute step advantages $\Phi_{step}(A_{i,t})$ via Equation~(\ref{eq:step_advantage})

    \State \small{\color{gray}{// Policy update phase}}
    \State Combine these advantages: $\Phi(A_{i,t}) = \Phi_{step}(A_{i,t})\cdot \Phi_{traj}(\tau_i)$
    \State Update policy $\theta$ by maximizing objective same as GRPO
\EndFor
\end{algorithmic}
\label{tab:pseudo_code}
\end{algorithm*}


\begin{table*}[t]
\centering
\resizebox{0.7\textwidth}{!}{
\begin{tabular}{lcccccccc}
\toprule
\multirow{2}{*}[ -3pt ]{Method} & \multirow{2}{*}[ -3pt ]{Window Size} & \multicolumn{7}{c}{\textbf{ALFWorld}} \\
\cmidrule(lr){3-9}
& & Pick & Cool & Heat & Look & Pick2 & Clean & All \\
\midrule
\multirow{4}{*}[-10pt]{GiGPO} 
& 2   & 80.0 & 72.0 & \underline{87.5} & \underline{84.6} & \underline{79.2} & 74.1 & 78.6 \\
& 4   & \textbf{91.4} & \textbf{84.0} & \textbf{93.8} & \underline{84.6} & \textbf{95.8} & \underline{92.6} & \textbf{90.7} \\
& 6   & 88.6 & \textbf{84.0} & \textbf{93.8} & \textbf{92.3} & \underline{79.2} & \underline{92.6} & \underline{87.9} \\
& 8   & 74.3 & \underline{80.0} & \underline{87.5} & \textbf{92.3} & \underline{79.2} & \underline{92.6} & 82.9 \\
& 10  & 80.0 & \underline{80.0} & \textbf{93.8} & 76.9 & 75.0 & \underline{92.6} & 82.9 \\
& 12  & \underline{88.6} & \underline{80.0} & \textbf{93.8} & 76.9 & 70.8 & \textbf{96.3} & 85.0 \\
\bottomrule
\end{tabular}
} 
\caption{Performance of GiGPO with different context window sizes on ALFWorld. \textbf{Bold scores} indicate the best performance, while \underline{underlined scores} represent the second-best.}
\label{tab:varing_size_gigpo}
\end{table*}

\begin{figure*}[t]
\begin{tcolorbox}[
    colback=gray!5,
    colframe=gray!150,
    coltitle=white,
    title=A Case for Efficient Search via Parallel Actions,
    fonttitle=\bfseries,
    boxrule=1.5pt,
    top=4pt, bottom=4pt, left=8pt, right=8pt,
    width=\textwidth, 
]
\textbf{User:}

Your task is to: put some candle on toilet.
Your current observation is: -= Welcome to TextWorld, ALFRED! =-

You are in the middle of a room. Looking quickly around you, you see a bathtubbasin 1, a garbagecan 1, a handtowelholder 1, a shelf 3, a shelf 2, a shelf 1, a sinkbasin 1, a toilet 1, a toiletpaperhanger 1, and a towelholder 1. 

Your admissible actions of the current situation are: ['go to bathtubbasin 1', 'go to garbagecan 1', 'go to handtowelholder 1', 'go to shelf 1', 'go to shelf 2', 'go to shelf 3', 'go to sinkbasin 1', 'go to toilet 1', 'go to toiletpaperhanger 1', 'go to towelholder 1', 'help', 'inventory', 'look'].

\textbf{Assistant:}
\begin{verbatim}
<think>
\end{verbatim}

My task is to put some candle on toilet. First, I need to find a candle. Looking at the objects in the room, I don't see a candle mentioned in the current observation. The objects are: bathtubbasin 1, garbagecan 1, handtowelholder 1, shelf 3, shelf 2, shelf 1, sinkbasin 1, toilet 1, toiletpaperhanger 1, and towelholder 1.

Candles are often found on shelves or in other storage areas. I should check the shelves first since they are the most likely places to contain items like candles. There are three shelves: shelf 1, shelf 2, and shelf 3.

Since I can explore up to 5 different environments, I can check multiple shelves simultaneously to speed up the search for the candle. I'll start by going to shelf 1, shelf 2, and shelf 3 in parallel environments.

In each environment, after going to the shelf, I'll need to look around to see if there's a candle there. But for now, I'll just navigate to the shelves first.
\begin{verbatim}
</think> 
<parallel>
<env_1> go to shelf 1 </env_1>
<env_2> go to shelf 2 </env_2>
<env_3> go to shelf 3 </env_3>
</parallel>
\end{verbatim}
\end{tcolorbox}
\caption{A Case for Efficient Search via Parallel Actions.}
\label{fig:parallel_explore}
\end{figure*}

\paragraph{Validating Hypotheses via Parallel Exploration} As the case shown in Figure~\ref{fig:validate_via_paralllel}, when the agent repeatedly fails to take book 2 from the bed, it hypothesizes that there might be a carrying limit. To test this, it uses parallel environments to check its inventory in env\_1 and env\_2 while also looking around in env\_3 to gather more context. This demonstrates the model’s ability to perform reasoning and efficiently validate hypotheses through parallel exploration.
\begin{figure*}[t]
\begin{tcolorbox}[
    colback=gray!5,
    colframe=gray!150,
    coltitle=white,
    title=A Case for Validating Hypotheses via Parallel Exploration,
    fonttitle=\bfseries,
    boxrule=1.5pt,
    top=4pt, bottom=4pt, left=8pt, right=8pt,
    width=\textwidth, 
]
\textbf{User:}
You are an expert agent operating in the ALFRED Embodied Environment. 

Your task is to: put two book in desk.

Your initial observation is: -= Welcome to TextWorld, ALFRED! =-

You are in the middle of a room. Looking quickly around you, you see a bed 1, a desk 1, a drawer 6, a drawer 5, a drawer 4, a drawer 3, a drawer 2, a drawer 1, a garbagecan 1, a shelf 2, a shelf 1, and a sidetable 1..
You have already taken multiple actions in multiple parallel environments. Below are the most recent observations and the corresponding actions you took:

...

\textcolor{red}{In the past few steps, the agent successfully picked up book 1 in environment 2, but has repeatedly failed to take book 2 from the same bed.}

\textbf{Assistant:}
\begin{verbatim}
<think>
\end{verbatim}
It seems like I am having trouble taking book 2 from bed 1 in environment 2. The action "take book 2 from bed 1" keeps resulting in "Nothing happens." This might be because I can only carry one item at a time, or there might be another issue.

Let me check the inventory in environment 2 to see what I\'m currently carrying. I should also check if there are any other books available in the room that I could use as alternatives.

Since I already have book 1, I should first put it down somewhere, then go back to get book 2. Or maybe I need to go to the desk with book 1 first, then come back for book 2.

Let me check the inventory in environment 2 first to understand the current state.
\begin{verbatim}
</think> 
<parallel>
<env_1> inventory </env_1>
<env_2> inventory </env_2>
<env_3> look </env_3>
</parallel>
\end{verbatim}
\end{tcolorbox}
\caption{A Case for Validating Hypotheses via Parallel Exploration.}
\label{fig:validate_via_paralllel}
\end{figure*}

\begin{figure*}[t]
\begin{tcolorbox}[
    colback=gray!5,
    colframe=gray!150,
    coltitle=white,
    title=A Case for Keep Exploring Even After Finding a Solution,
    fonttitle=\bfseries,
    boxrule=1.5pt,
    top=4pt, bottom=4pt, left=8pt, right=8pt,
    width=\textwidth, 
]
\textbf{User:}
You are an expert agent operating in the ALFRED Embodied Environment. 

Your task is to: put a cool mug in coffeemachine.

Your initial observation is: -= Welcome to TextWorld, ALFRED! =-

You are in the middle of a room. Looking quickly around you, you see a cabinet 27, a cabinet 26, a cabinet 25, a cabinet 24, a cabinet 23, a cabinet 22, a cabinet 21, a cabinet 20, a cabinet 19, a cabinet 18, a cabinet 17, a cabinet 16, a cabinet 15, a cabinet 14, a cabinet 13, a cabinet 12, a cabinet 11, a cabinet 10, a cabinet 9, a cabinet 8, a cabinet 7, a cabinet 6, a cabinet 5, a cabinet 4, a cabinet 3, a cabinet 2, a cabinet 1, a coffeemachine 1, a countertop 2, a countertop 1, a diningtable 1, a drawer 12, a drawer 11, a drawer 10, a drawer 9, a drawer 8, a drawer 7, a drawer 6, a drawer 5, a drawer 4, a drawer 3, a drawer 2, a drawer 1, a fridge 1, a garbagecan 1, a microwave 1, a sinkbasin 1, a stoveburner 4, a stoveburner 3, a stoveburner 2, a stoveburner 1, and a toaster 1.
In your last step, your actions, corresponding observations and admissible actions are: 

...

\textbf{Assistant:}
\begin{verbatim}
<think>
\end{verbatim}
Perfect! I\'m still at countertop 2 in environment 4 where I found the mug. Now I need to take the mug and then put it in the coffeemachine. The task requires a "cool mug" though, so I need to cool it down first. I should take the mug and then go to the fridge to cool it.

Let me take the mug from countertop 2 in environment 4, and continue exploring other environments in parallel.
\begin{verbatim}
</think> 
<parallel>
<env_1>open cabinet 2</env_1>
<env_2>open cabinet 11</env_2>
<env_3>go to drawer 5</env_3>
<env_4>take mug 1 from countertop 2</env_4>
<env_5>open cabinet 21</env_5>
</parallel>
\end{verbatim}
\end{tcolorbox}
\caption{A Case for Keep Exploring Even After Finding a Solution.}
\label{fig:keep_explore}
\end{figure*}

\begin{figure*}[t]
\begin{tcolorbox}[
    colback=gray!5,
    colframe=gray!150,
    coltitle=white,
    title=A Case for Structured Search Across Location Types,
    fonttitle=\bfseries,
    boxrule=1.5pt,
    top=4pt, bottom=4pt, left=8pt, right=8pt,
    width=\textwidth, 
]
\textbf{User:}
You are an expert agent operating in the ALFRED Embodied Environment. 

Your task is to: cool some tomato and put it in microwave.

Your initial observation is: -= Welcome to TextWorld, ALFRED! =-

You are in the middle of a room. Looking quickly around you, you see a cabinet 26, a cabinet 25, a cabinet 24, a cabinet 23, a cabinet 22, a cabinet 21, a cabinet 20, a cabinet 19, a cabinet 18, a cabinet 17, a cabinet 16, a cabinet 15, a cabinet 14, a cabinet 13, a cabinet 12, a cabinet 11, a cabinet 10, a cabinet 9, a cabinet 8, a cabinet 7, a cabinet 6, a cabinet 5, a cabinet 4, a cabinet 3, a cabinet 2, a cabinet 1, a coffeemachine 1, a countertop 3, a countertop 2, a countertop 1, a drawer 12, a drawer 11, a drawer 10, a drawer 9, a drawer 8, a drawer 7, a drawer 6, a drawer 5, a drawer 4, a drawer 3, a drawer 2, a drawer 1, a fridge 1, a garbagecan 1, a microwave 1, a sinkbasin 1, a stoveburner 4, a stoveburner 3, a stoveburner 2, a stoveburner 1, and a toaster 1.

...

\textcolor{red}{The agent went to and opened the fridge in all 5 environments, but found it empty, no tomato located yet.}

\textbf{Assistant:}
\begin{verbatim}
<think>
\end{verbatim}
The fridge is empty in all environments. I need to search other locations for a tomato. Tomatoes might be in cabinets, drawers, or on countertops. I should search systematically. Since I have multiple environments, I can search different locations simultaneously.

I'll use the environments to search different types of locations:
- Environment 1: Search a cabinet (e.g., cabinet 1)

- Environment 2: Search a drawer (e.g., drawer 1)

- Environment 3: Search countertop 1

- Environment 4: Search countertop 2

- Environment 5: Search countertop 3

This will help me find the tomato faster by searching multiple locations at once.

\begin{verbatim}
</think> 
<parallel>
<env_1> go to cabinet 1 </env_1>
<env_2> go to drawer 1 </env_2>
<env_3> go to countertop 1 </env_3>
<env_4> go to countertop 2 </env_4>
<env_5> go to countertop 3 </env_5>
</parallel>
\end{verbatim}
\end{tcolorbox}
\caption{A Case for Structured Search Across Location Types.}
\label{fig:structure_explore}
\end{figure*}

\end{document}